%% file: main.tex
\documentclass[fleqn,10pt]{wlscirep}
\usepackage[utf8]{inputenc}
\usepackage[T1]{fontenc}
\usepackage{multirow}
\input{abstract} 
\begin{document}
\title{Human locomotor control timescales depend on the environmental context and sensory input modality}
\author[1]{Wei-Chen Wang}
\author[2]{Antoine De Comite}
\author[4,6]{Alexandra Voloshina}
\author[4,5,6]{Monica Daley}
\author[1,2,3,*]{Nidhi Seethapathi}
\affil[1]{Department of Electrical Engineering and Computer Science, School of Engineering, MIT, Cambridge, MA 02139, USA.}
\affil[2]{McGovern Institute for Brain Research, MIT, Cambridge, MA 02139, USA.}
\affil[3]{Brain and Cognitive Sciences, School of Science, MIT, Cambridge, MA 02139, USA.}
\affil[4]{Mechanical and Aerospace Engineering, Henry Samueli School of Engineering, UC Irvine, Irvine CA 92697, USA.}
\affil[5]{Ecology and Evolutionary Biology, School of Biological Sciences, UC Irvine, Irvine CA 92697, USA. } 
\affil[6]{Biomedical Engineering, Henry Samueli School of Engineering, UC Irvine, Irvine CA 92697, USA. } 
\affil[*]{nidhise@mit.edu}

\flushbottom
\maketitle
\thispagestyle{empty}

\input{introduction} 
\input{results} 
\input{discussion}
\input{method} 
\input{acknowledgements} 
\bibliography{main}
\clearpage 
\appendix 
\input{appendices}
\end{document}

%% file: abstract.tex
\begin{abstract}
Everyday locomotion is a complex sensorimotor process that can unfold over multiple timescales, from long-term path planning to rapid, reactive adjustments. However, we lack an understanding of how factors such as environmental demands, or the available sensory information simultaneously influence these control timescales. To address this, we present a unified data-driven framework to quantify the control timescales by identifying how early we can predict future actions from past inputs.  We apply this framework across tasks including walking and running, environmental contexts including treadmill, overground, and varied terrains, and sensory input modalities including gaze fixations and body states. We find that deep neural network architectures that effectively handle long-range dependencies, specifically Gated Recurrent Units and Transformers, outperform other architectures and widely used linear models when predicting future actions. Our framework reveals the factors that influence locomotor foot placement control timescales. Across environmental contexts, we discover that humans rely more on fast timescale control in more complex terrain. Across input modalities, we find a hierarchy of control timescales where gaze predicts foot placement before full-body states, which predict before center-of-mass states. Our model also identifies mid-swing as a critical phase when the swing foot's state predicts its future placement, with this timescale adapting across environments. Overall, this work offers data-driven insights into locomotor control in everyday settings, offering models that can be integrated with rehabilitation technologies and movement simulations to improve their applicability in everyday settings.
\end{abstract} 

%% file: introduction.tex
%\linenumbers
\section*{Introduction}
Understanding the mechanisms humans use to control everyday movements is a fundamental scientific challenge for sensorimotor control, neuromechanics, and rehabilitation engineering. For instance, a critical aspect of locomotor stability is the control of foot placement relative to the body \cite{townsend1985biped, bruijn2018control}; the timescale on which this control unfolds can vary from faster reactive control \cite{rankin2014neuromechanical, horslen2014modulation} to slower proactive selection of footholds \cite{matthis2018gaze, muller2024foothold}. Moreover, the existence of multiple control timescales likely maps to distinct sensorimotor and neural underpinnings \cite{shadmehr2008computational, richer2024mobile}, and may serve different environment-dependent functional goals \cite{darici2023humans, seethapathi2024exploration}. To achieve adaptive locomotion, humans must therefore modulate control across multiple timescales informed by the environmental context and the available sensory information. Despite advances in collecting high-throughput movement data in real-world settings, we lack an understanding of how multiple control timescales are chosen and how they depend on the environmental and sensory context. This study introduces a data-driven framework to characterize multi-timescale control strategies in locomotion across diverse environmental settings and sensory input modalities. 

\begin{figure}[tp] 
    \centering
    \includegraphics[width=\linewidth]{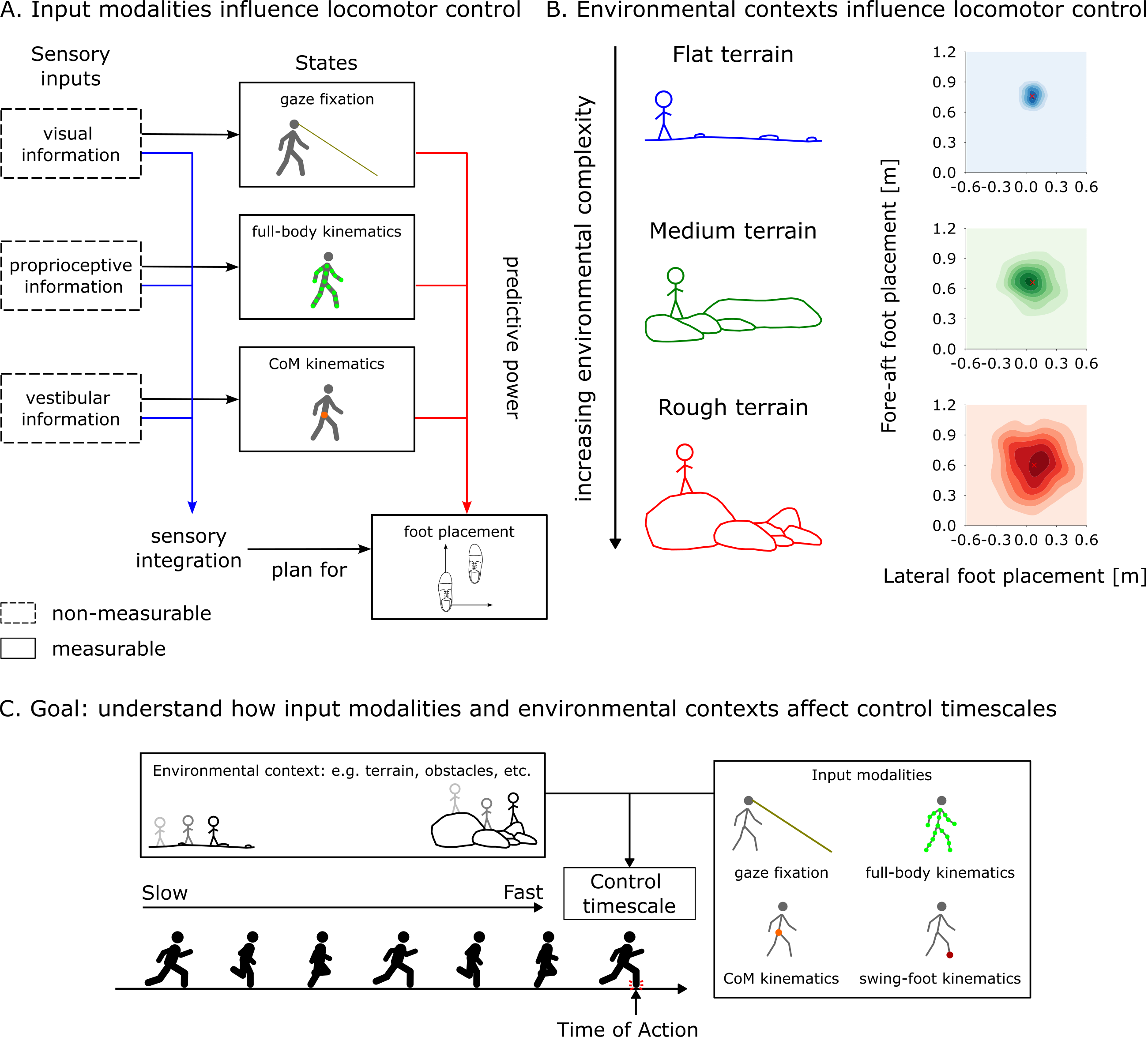}
    \caption{
    \textbf{A} Humans integrate sensory inputs to plan for future foot placements during locomotion. While sensory inputs such as visual, proprioceptive and vestibular information are not readily measurable in humans, we use gaze fixation, full-body and center-of-mass kinematics to indirectly infer their predictive influence on foot placement. \textbf{B} We analyze walking data across varied terrains to explore how humans adapt foot placement control strategies in response to increasing environmental complexity. The heatmaps illustrate lateral and fore-aft right foot placement relative to the left foot across terrains, with gradients derived from kernel density estimation. \textbf{C} We seek to quantifies the control timescale under diverse environmental contexts, and across various input modalities, including gaze fixation, full-body, center-of-mass, and swing foot kinematics.} 
    \label{fig:expo}
\end{figure}

Using data-driven models to understand multi-timescale control strategies in complex human behaviors necessitates some form of ``system identification'' i.e., mapping future actions to a history of time-varying inputs. While early data-driven models for system identification of locomotor control assume linear and fixed timescale mappings between states and actions  \cite{wang2014stepping, seethapathi2019step, afschrift2021similar}, it is unclear whether these assumptions generalize to more complex settings. Data-driven models that leverage deep learning architectures offer a promising approach for capturing such context-dependent and nonlinear dynamics \cite{radosavovic2025humanoid}. However, prior attempts to develop such nonlinear data-driven models of locomotor control have not been used to understand the underlying control strategies and have been limited to a single network architecture \cite{ xiong2023probability, chen2021probability}, a single input modality \cite{lee2023deep, xiong2023probability}, or to laboratory-constrained datasets \cite{chen2021probability, asogwa2022using}. To overcome these limitations, we put forward a generalizable data-driven framework to understand locomotor control in real-world settings, testing its predictive performance across different network architectures, different input modalities (e.g. gaze, full-body kinematics), and multiple environmental contexts (e.g. varied terrains). Using these models, we provide insight into how the environmental context and sensory input modality influence multi-timescale control.

Biological motor control integrates information from multiple sensory input modalities to plan future actions \cite{wolpert2000computational, van1999integration}. During locomotion, foot placement is influenced by vision \cite{matthis2014visual, matthis2018gaze}, center-of-mass (CoM) states \cite{wang2014stepping, seethapathi2019step}, and postural information \cite{dietz2002proprioception, roden2015hip}.  While direct measurement of the underlying sensory inputs (e.g. visual, proprioceptive, and vestibular signals) is typically impossible in humans, we can investigate their influence by studying measurable proxies like gaze fixations, full-body states, and  CoM states (Figure \ref{fig:expo}A). Existing data-driven models of locomotion often oversimplify the influence of multiple sensory input modalities by focusing on a single input and assuming a fixed control timescale \cite{wang2014stepping, afschrift2021similar}. In this study, we systematically evaluate how the timescale of foot placement control is influenced by various input modalities by characterizing when information from a given input becomes available for prediction (Figure \ref{fig:expo}C). By characterizing how the control timescales depend on the sensory input modality, we reveal critical phases within the movement when each input becomes useful for control.

Biological motor control is continuously tuned to environment-specific demands, employing adaptive strategies \cite{vetter2000context}. To maintain stability, individuals adjust their actions in response to changing environmental features like varying terrains or obstacles \cite{hayhoe2018control, patla1997understanding}. As the environmental complexity increases, humans may adjust their sensorimotor control strategies  (Figure \ref{fig:expo}B); perhaps by relying more on visual gaze  \cite{matthis2017critical, hayhoe2018control} than on other sensory input modalities, thereby explicitly prioritizing safer footholds over more automatic foot placement choices. These adjustments can occur over multiple timescales, ranging from within-step corrections to longer-term strategic planning over several steps \cite{patla1997understanding, warren2006dynamics, muller2024foothold}. The possibility of multiple control timescales could also point to distinct underlying neural computations \cite{shadmehr2008computational} or distinct functional goals \cite{darici2023humans, seethapathi2024exploration}. It is therefore a necessary first step to quantify these control timescales to advance our scientific understanding of motor control in real-world settings. Here, we use the relative predictive power of data-driven models trained on distinct input modalities to identify these control timescales, thereby revealing how they depend on the environmental context. Specifically, we discover an environmental context-dependent shift from slow to fast control timescales in human locomotion. 

In this study, we (i) develop data-driven models, leveraging deep learning architectures, to predict future foot placements from past input states, and (ii) utilize these models to understand how the timescale of locomotor control depends on the environmental context and the sensory input modality. We found that neural network architectures suited to capturing long-range dependencies generalize best across different contexts. Our data-driven models reveal that the locomotor control timescale depends on both the environmental context and the sensory input modality, highlighting when different inputs become useful for control and how the control timescale shifts with environmental demands. Collectively, this work offers data-driven insights into the influence of environmental context and sensory information on locomotor control, and these data-driven models can be extended to characterize sensorimotor disorders, design human-aware wearable assistive technologies, and develop more human-like simulations of complex everyday human movements.

%% file: results.tex
\section*{Results} 
\begin{figure}[htp] 
    \centering
    \includegraphics[width=\linewidth]{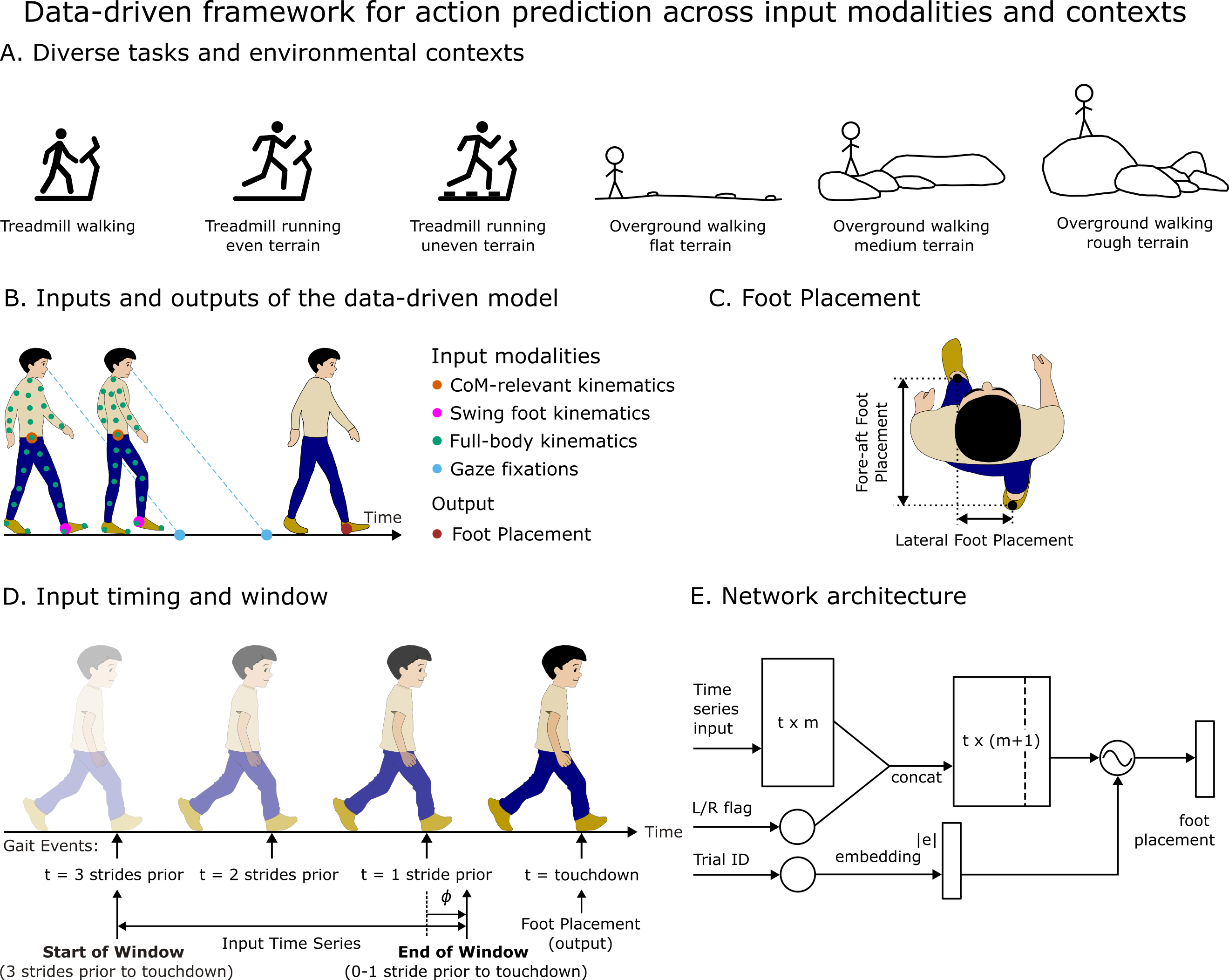}
    \caption{
    \textbf{Overview of the data-driven framework.}
    \textbf{A} We include diverse tasks and contexts, including overground walking with various terrain conditions, treadmill walking, and treadmill running on even and uneven terrain. 
    \textbf{B} Inputs contain hypothesized modalities such as CoM-relevant kinematics, swing foot kinematics, full-body kinematics, and gaze fixations (defined by the intersection of the gaze vector with the ground plane). The output is foot placement. 
    \textbf{C} We define foot placement output as the lateral and fore-aft foot placement relative to the opposite foot at its previous heel-strike.  
    \textbf{D} We use a time series spanning from three strides prior to touchdown to a specific gait phase within the previous stride. The output corresponds to foot placement at heel-strike. We denote the ending gait phase of the time series relative to one stride prior to touchdown as $\phi$, discretized and sampled from $n=21$ equally spaced phases within $0 \leq \phi \leq 1$.  
    \textbf{E} The network architecture of the nonlinear models. The input consists of a time series, an L/R flag, and a trial ID. The time series and L/R flag are concatenated, while the trial ID is embedded to help the model learn trial-specific information. We incorporate this embedding in an architecture-dependent manner, i.e. it influences the model prediction differently depending on the neural network architecture, as detailed in the Methods section.}
    \label{fig:fig1}
\end{figure}

In this section, we present findings developing a data-driven framework to predict future actions from past input states and using these models to understand how the environment and the input modality influence locomotor foot placement control timescales. First, we find that deep learning architectures, particularly GRUs and Transformers, significantly outperform other architectures and widely used linear models across different environmental settings. Using these models, we find that the degree to which humans rely on faster versus slower foot placement predictions during overground walking depends on the terrain complexity. Next, we discover that the swing foot input state becomes predictive of future foot placement in the middle of the swing phase, and how this timing changes across terrains. Finally, by comparing the predictive power of distinct input modalities to a baseline, we characterize how the timescale of prediction depends on the sensory input modality. An overview of the key methodological components of the data-driven framework can be found in Figure \ref{fig:fig1}.

\subsection*{Models with flexible nonlinear input dependence best predict future foot placement across settings}
We aimed to test which data-driven model architectures best predict future foot placement from a history of input states, comparing many deep neural network architectures and traditionally used linear models \cite{wang2014stepping, bruijn2013assessing} across input modalities and environmental contexts. Specifically, we evaluated the performance of nine models — GRU, Transformer, LSTM, FCNN, TCN, Linear instance with (LI2) and without (LI) L2 regularization, Linear history with (LH2) and without (LH) L2 regularization (see Methods for details) — based on their ability to predict foot placement across settings including treadmill walking, treadmill running, and overground walking on various terrains, and input modalities including CoM-relevant kinematics, full-body kinematics, swing foot kinematics, and gaze fixations. We consider the predictive power of the model using swing foot kinematics as the baseline model, which indicates how much information is contained in the foot itself over the gait cycle for predicting future foot placement. We evaluated each model's performance based on the Root Mean Squared Error (RMSE) of its predictions relative to that of the optimal model at each gait phase, and then normalized by the highest-scoring model. The Methods section details the baseline model and the evaluation metric. 

We find that during treadmill walking and running, the body state to foot placement mappings are effectively captured by linear models, particularly model LI2 which includes regularization (within 3\% of performance; Table \ref{table:model-performance}). However, the linear models are severely outperformed by nonlinear models in the overground locomotion tasks (5-13\% worse for CoM as input modality, 13-39\% worse for other input modalities; Table \ref{table:model-performance}). Nonlinear models are therefore essential for capturing the history-dependence and inherent nonlinearities that map input states to foot placements during non-stationary activities, such as walking overground. Among the nonlinear architectures, GRU demonstrates the best overall performance across tasks, followed closely by Transformer and LSTM architectures. An exception arises when using gaze fixations as the input modality during flat terrain overground walking, where Transformers significantly outperform all other models. However, this is also the only input modality and context where the predictive power of the model with the gaze as input to predict future foot placement never surpasses the baseline model which uses the foot state to predict future foot placement, suggesting that gaze does not play a significant role in foot placement prediction during overground walking on flat terrain. To examine the temporal trends of each model more closely, we provide the RMSE as a function of gait phases for each model under different conditions in Figures \ref{fig:fig6} and \ref{fig:fig7} in the Supplementary Information. We find that FCNN performs better during treadmill locomotion, but poorly during overground walking, whereas TCN performs poorly during treadmill locomotion, even worse than the optimal linear model, but better during overground walking. Since GRU demonstrates the best overall performance, we use it for the analyses in the subsequent sections of the Results. 

\input{table1}

\subsection*{Faster versus slower prediction timescales depend on the environmental context}
\begin{figure}[htp] 
    \centering
    \includegraphics[width=0.914\linewidth]{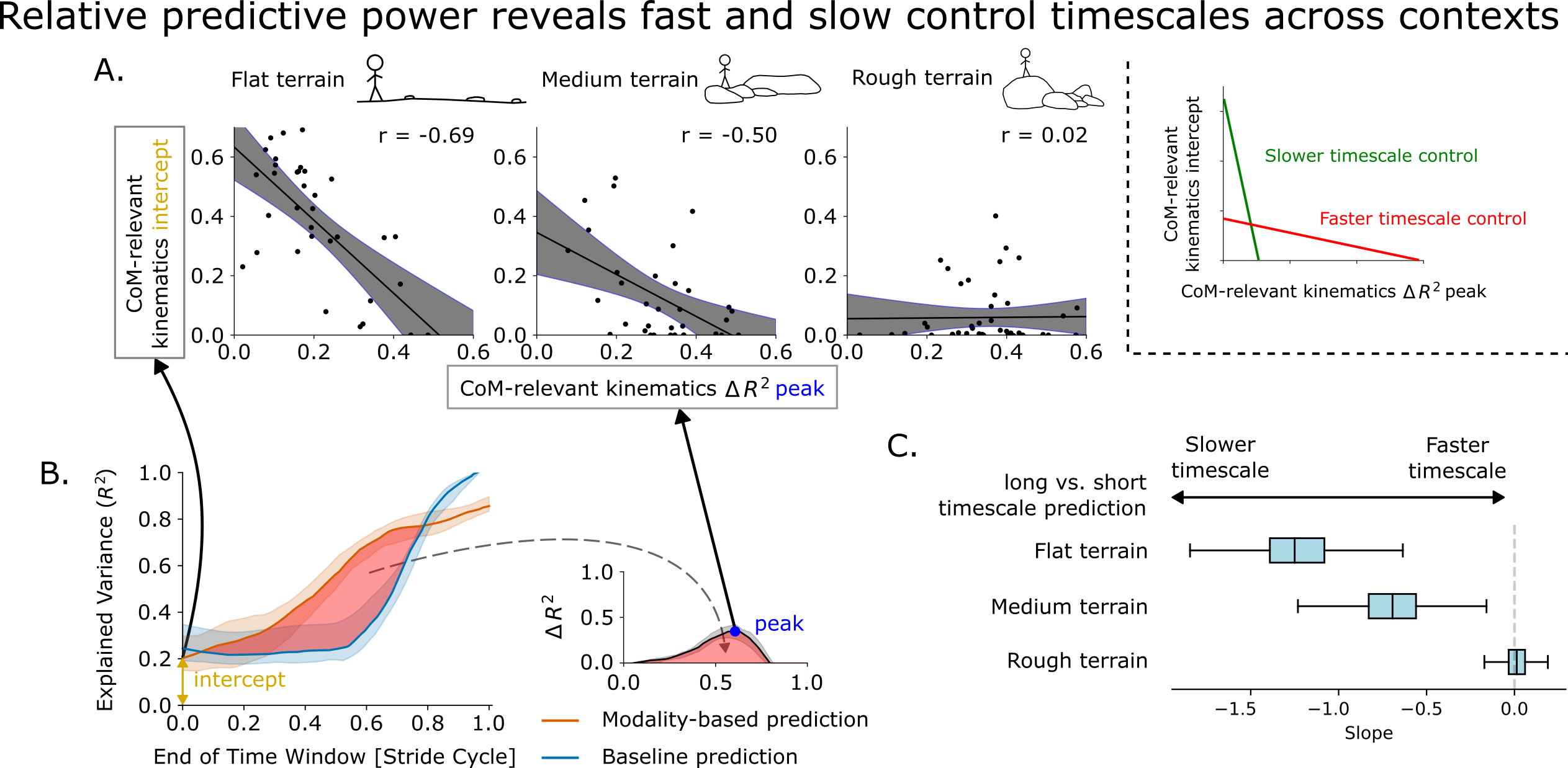}
    \caption{
    \textbf{Relative predictive power reveals fast and slow timescales across contexts. }
    \textbf{A} Relationship between the CoM-relevant kinematics intercepts and peak $\Delta R^2$ (both defined in \textbf{B}) during overground walking across terrains with linear regression fits. The plots display the regression line and 95\% confidence interval. 
    \textbf{B} Explained variance ($R^2$) as a function of gait phase for modality-based (orange) and autoregressive baseline (blue) predictions. The intercept is defined as the modality-based $R^2$ value at gait phase 0 (i.e., the input time series spans from three gait cycles prior to touchdown and ends one gait cycle before touchdown). The red shaded area represents the relative predictive power ($\Delta R^2$), calculated as the difference between modality-based and baseline predictions. The peak is the maximum $\Delta R^2$. 
    \textbf{C} Linear coefficients between CoM-relevant kinematics $\Delta R^2$ peak and intercept of CoM-relevant kinematics. The box-plot shows the median (bar in box), $25-75\%$ percentile (box), and non-outlier range (whiskers) of the bootstrap statistics. The magnitude of the slope quantifies the strength of the tradeoff between fast- and slow-timescale prediction. 
    }
    \label{fig:fig3}
\end{figure}

Biological motor control exhibits remarkable adaptability, seamlessly combining fast and slow control strategies to meet environmental demands. Here, we quantify these timescales by measuring how well an individual's body states predict their future foot placement across different environments (Figure~\ref{fig:expo}), using the data-driven framework from the previous section. For clarity, we first outline how the faster and slower timescales are computed (also shown in Figure~\ref{fig:fig3}B):
\begin{itemize}
    \item We define the \textbf{slower} prediction timescale through the \textbf{intercept} of a model trained on the CoM kinematics. This intercept is the $R^2$ of the model at gait phase 0, i.e. using an input window spanning from six to two steps before foot placement.
    \item We define the \textbf{faster} prediction timescale through the peak of the \textbf{relative predictive power} of the CoM-based model (Figure~\ref{fig:fig3}B). This relative power ($\Delta R^2$) is calculated by subtracting the predictive power of a baseline model (which predicts foot placement from past foot states) from the CoM-based model $R^2$. The peak is the maximum of this $\Delta R^2$.
\end{itemize}

While different trials show varying intercepts, the intercepts of the CoM-based kinematics and the baseline model are highly correlated ($r = 0.97$). Thus, subtracting the baseline predictive power from the CoM predictive power effectively reduces inter-trial variability, providing more interpretable comparisons across environmental contexts. We found a trade-off between these fast and slow control timescales during overground walking, revealed by a negative correlation that depends on the environment. This negative correlation indicates a greater reliance on either slow- or fast-timescale prediction depending on the terrain (Figure~\ref{fig:fig3}A, inset). As the terrain becomes rougher, individuals rely less on slow-timescale prediction, indicated by a decrease in the intercept of the CoM-based kinematics and a decrease in its correlation with the $\Delta R^2$ peak, from $r = -0.69$ on flat terrain to $r = -0.5$ on medium terrain, and $r = 0.02$ on rough terrain (Figure~\ref{fig:fig3}A). The change in the slope of the relationship between peak relative predictive power and the intercept across terrains suggests that the balance between slow- and fast-timescale foot placement predictions shifts with the environmental context (Figure~\ref{fig:fig3}C).

\subsection*{Swing foot kinematics predict future foot placement mid-swing} 
\begin{figure}[htp] 
    \centering
    \includegraphics[width=0.74\linewidth]{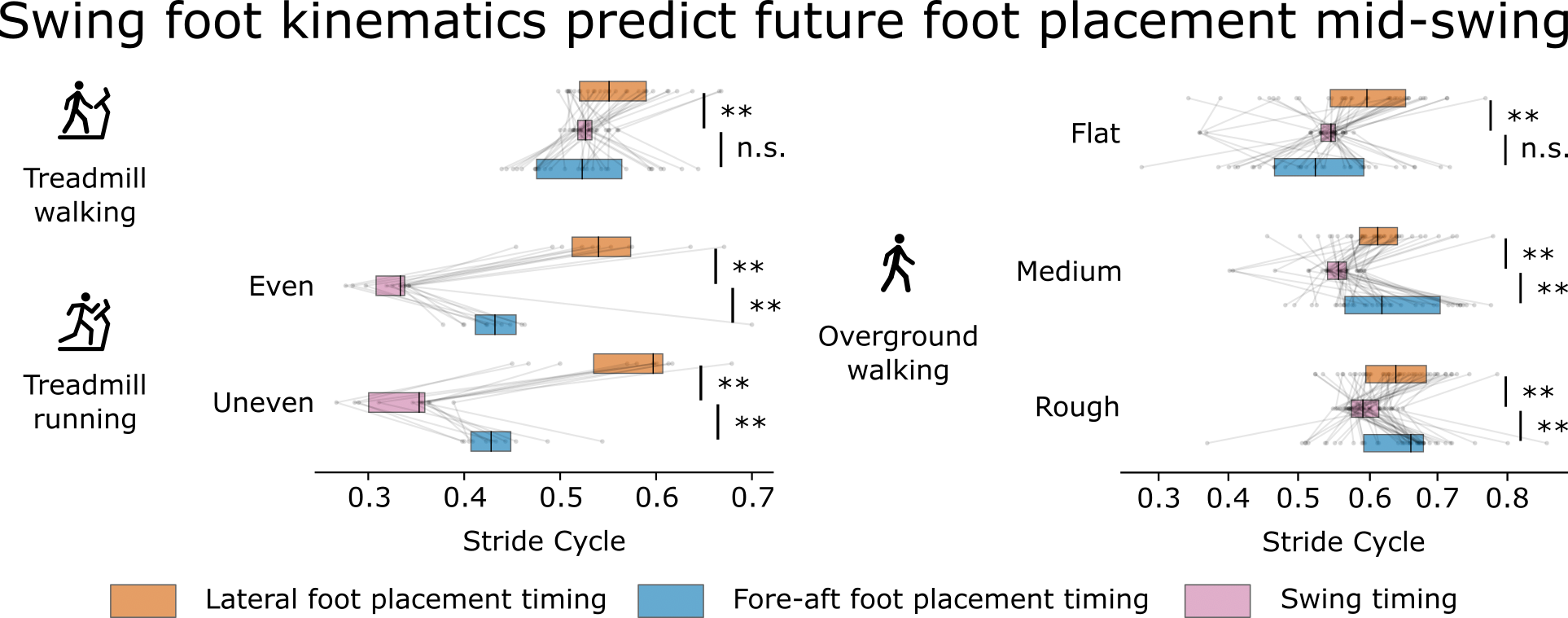}
    \caption{\textbf{Swing foot kinematics predict future foot placement mid-swing}. Comparison of the timing when the swing phase begins to when the swing foot begins to predict its future placement, detected using the methods outlined in Figure \ref{fig:fig2}. Comparisons are shown for all tasks and environmental contexts including treadmill walking, treadmill running and overground walking across terrain variations. $^{**}$ indicates $p<0.01$; $n.s.$ indicates not significant. 
    }
    \label{fig:fig5}
\end{figure}

Swing foot control is an essential strategy for achieving stable locomotion in varying terrains and tasks. In prior work, it is implicitly assumed that swing control during walking occurs simultaneously with the timing of swing initiation \cite{wang2014stepping, seethapathi2019step, joshi2019controller}. However, our data-driven framework reveals that swing foot initiation timing is not necessarily the same as when the swing foot begins to predict its future placement. We find that the timing when the swing foot begins to contain significant information to predict its future placement, namely the ``breakpoint'' timing, is different from the swing initiation timing. For the purpose of this comparison, the timing of swing initiation is defined as when the swing foot's velocity exceeds 5\% of its peak velocity. We find that, during treadmill and overground walking on flat terrain, the lateral foot placement prediction (FP) timing occurs significantly later than swing foot initiation (Wilcoxon signed-rank test, one-sided, $p<0.05$; Figure \ref{fig:fig5}). However, we do not find this to be true for fore-aft FP timing. On the other hand, for overground walking on uneven terrain, both the lateral and fore-aft FP timing were delayed relative to swing initiation (Figure \ref{fig:fig5}). Similarly, during treadmill running, where swing initiation occurs earlier due to the existence of a flight phase, both lateral and fore-aft FP timing occur after swing initiation (Figure \ref{fig:fig5}). This further highlights that while the swing foot begins its motion early, it need not contribute information about foot placement until later in the gait cycle \cite{hasaneini2014swing, hasaneini2013optimal}. Taken together, our findings reveal how swing foot kinematics are adapted to environmental and task-specific demands, and that swing initiation does not necessarily coincide with the point at which the swing foot begins to predict its future placement. This delay could reflect the additional planning demands of navigating complex environments.

\subsection*{Relative predictive power reveals different control timescales across input modalities} 
\begin{figure}[htp] 
    \centering
    \includegraphics[width=0.96\linewidth]{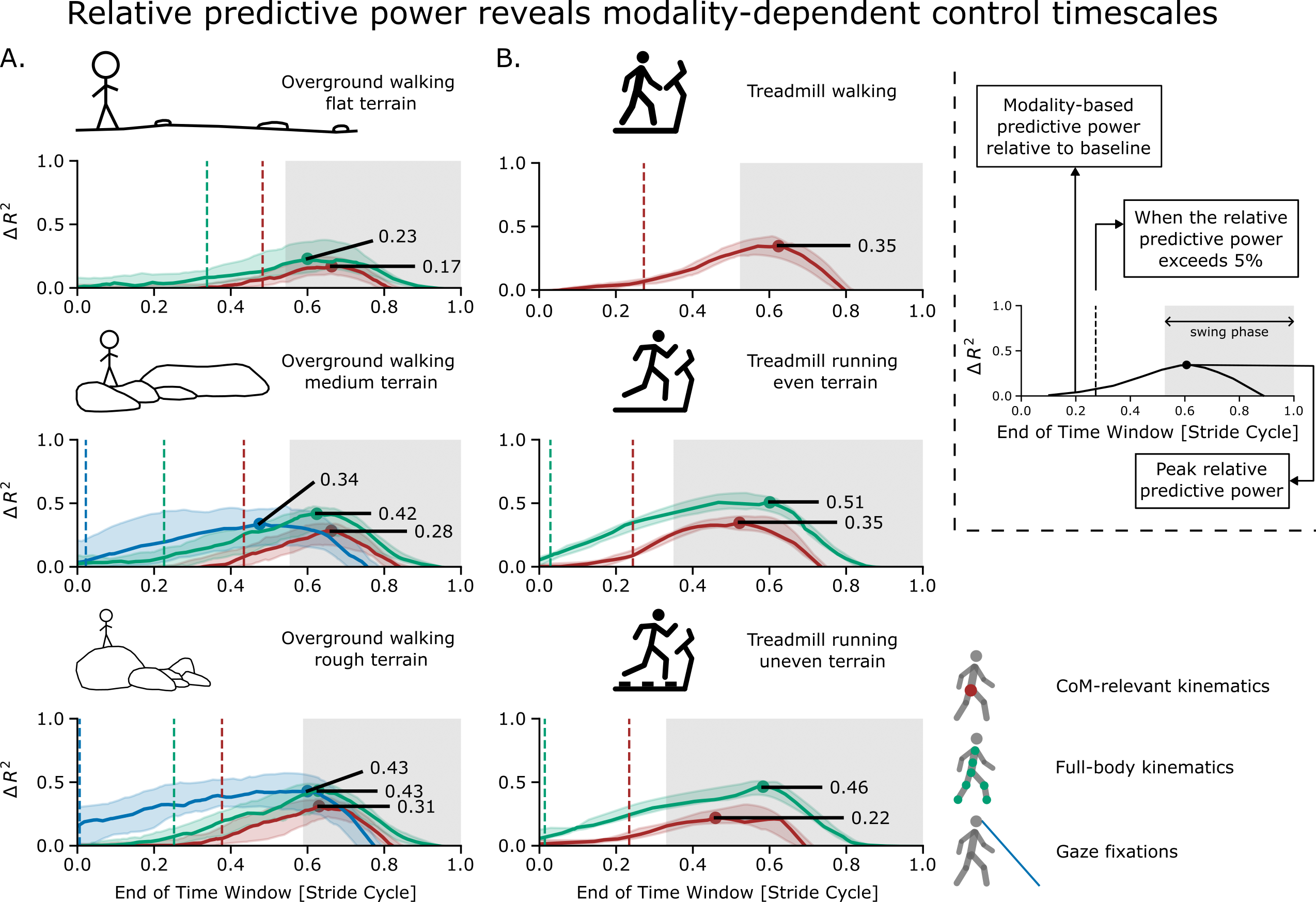}
    \caption{
    \textbf{Relative predictive power reveals modality-dependent control timescales. }
    \textbf{A} Relative predictive power ($\Delta\,R^2$, median and interquartile range) of CoM-relevant kinematics, full-body kinematics, and gaze fixations for overground walking on different terrain roughness. The highlighted interval begins when $\Delta\,R^2$ exceeds 5\% (Wilcoxon, one-sided), and ends at the peak $\Delta\,R^2$. The shaded area represents the swing phase. 
    \textbf{B} Relative predictive power of CoM-relevant kinematics for treadmill walking, and CoM-relevant kinematics and full-body kinematics for treadmill running on even and uneven terrains. 
    }
    \label{fig:fig4} 
\end{figure}

Human locomotion requires the dynamic integration of multiple sensory inputs to guide control. Here, we quantify the timescales at which different input modalities are used for foot placement control, including center of mass (CoM)-relevant kinematics, full-body kinematics, and gaze fixations. To do this, we trained our data-driven framework to predict future foot placement based on time series data from each modality. As the input time window increases, the predictive power of our models also increases (Figure \ref{fig:fig4}). We quantified the control timescale for each input by analyzing its relative predictive power ($\Delta R^2$) i.e. the improvement over a baseline model trained only on the foot state itself (Figure \ref{fig:fig4}, inset).

We found that full-body kinematics began to outperform the baseline model earlier than CoM-relevant kinematics. Despite including significantly more body markers (29 vs. 1), full-body kinematics only explained an additional 6--14\% of foot placement variance. This highlights the primary role of CoM-relevant kinematics in locomotor control (Figure \ref{fig:fig4}). During walking on more complex terrain, gaze was a more accurate predictor of foot placements than body states. Gaze-based predictions occurred earlier than both CoM-relevant and full-body kinematics for lateral (Figure \ref{fig:fig4}A) but not for fore-aft foot placements (Figure \ref{fig:fig8}C). This indicates that in challenging environments, humans use visual information to inform lateral foot placement earlier than information from the body states. Across environmental contexts and input modalities, lateral foot placement prediction timescales were earlier than the fore-aft timescales (Figure \ref{fig:fig8}A). This suggests a prioritization of lateral stability during locomotion, which is consistent with previous findings \cite{wang2014stepping, mcgeer1990passive, collins2008dynamic}. In the case of treadmill running, the timescale of lateral foot placement prediction was not significantly different between even and uneven terrain. However, predictions for fore-aft foot placement occurred earlier on uneven terrain (Figure \ref{fig:fig8}B). This discrepancy is likely because the uneven terrain running data only included fore-aft perturbations \cite{voloshina2015biomechanics}.

%% file: table1.tex
\begin{table}[htp]
\centering
\begin{tabular}{|c|c|c|c|c|c|c|c|c|c|c|c|} 
\midrule Task                                 & Terrain                  & M & GRU                         & Transformer                          & LSTM                         & FCNN                          & TCN                        & LI                           & LH                           & LI2                          & LH2                          \\ \midrule 
TW                    & N/A                      & C & \cellcolor[HTML]{32CB00}1.00 & \cellcolor[HTML]{9AFF99}0.98 & \cellcolor[HTML]{32CB00}1.00 & \cellcolor[HTML]{32CB00}0.98 & \cellcolor[HTML]{FD6864}0.71 & \cellcolor[HTML]{9AFF99}0.97 & \cellcolor[HTML]{FD6864}0.26 & \cellcolor[HTML]{9AFF99}0.97 & \cellcolor[HTML]{32CB00}0.99 \\ \midrule
                                     &                          & C & \cellcolor[HTML]{32CB00}1.00 & \cellcolor[HTML]{32CB00}0.98 & \cellcolor[HTML]{32CB00}0.98 & \cellcolor[HTML]{9AFF99}0.96 & \cellcolor[HTML]{FFCE93}0.92 & \cellcolor[HTML]{32CB00}1.00 & \cellcolor[HTML]{FD6864}0.34 & \cellcolor[HTML]{32CB00}1.00 & \cellcolor[HTML]{9AFF99}0.97 \\ \cmidrule{3-12} 
                                     & \multirow{-2}{*}{Even}   & F & \cellcolor[HTML]{32CB00}1.00 & \cellcolor[HTML]{FFCE93}0.94 & \cellcolor[HTML]{32CB00}1.00 & \cellcolor[HTML]{FD6864}0.89 & \cellcolor[HTML]{FD6864}0.82 & \cellcolor[HTML]{FD6864}0.38 & \cellcolor[HTML]{FD6864}0.62 & \cellcolor[HTML]{32CB00}1.00 & \cellcolor[HTML]{FD6864}0.78 \\ \cmidrule{2-12} 
                                     &                          & C & \cellcolor[HTML]{32CB00}1.00 & \cellcolor[HTML]{9AFF99}0.97 & \cellcolor[HTML]{32CB00}0.98 & \cellcolor[HTML]{9AFF99}0.96 & \cellcolor[HTML]{FFCE93}0.91 & \cellcolor[HTML]{9AFF99}0.97 & \cellcolor[HTML]{FD6864}0.31 & \cellcolor[HTML]{9AFF99}0.97 & \cellcolor[HTML]{9AFF99}0.97 \\ \cmidrule{3-12} 
\multirow{-4}{*}{TR}  & \multirow{-2}{*}{Uneven} & F & \cellcolor[HTML]{32CB00}1.00 & \cellcolor[HTML]{9AFF99}0.96 & \cellcolor[HTML]{32CB00}1.00 & \cellcolor[HTML]{FFCE93}0.93 & \cellcolor[HTML]{FD6864}0.91 & \cellcolor[HTML]{FD6864}0.35 & \cellcolor[HTML]{FD6864}0.63 & \cellcolor[HTML]{32CB00}0.99 & \cellcolor[HTML]{FD6864}0.76 \\ \midrule
                                     &                          & C & \cellcolor[HTML]{32CB00}1.00 & \cellcolor[HTML]{32CB00}0.99 & \cellcolor[HTML]{9AFF99}0.96 & \cellcolor[HTML]{9AFF99}0.95 & \cellcolor[HTML]{FFCE93}0.94 & \cellcolor[HTML]{FFCE93}0.91 & \cellcolor[HTML]{FD6864}0.70 & \cellcolor[HTML]{FFCE93}0.95 & \cellcolor[HTML]{FD6864}0.88 \\ \cmidrule{3-12} 
                                     &                          & F & \cellcolor[HTML]{32CB00}0.99 & \cellcolor[HTML]{32CB00}1.00 & \cellcolor[HTML]{9AFF99}0.97 & \cellcolor[HTML]{FFCE93}0.94 & \cellcolor[HTML]{FFCE93}0.95 & \cellcolor[HTML]{FD6864}0.52 & \cellcolor[HTML]{FD6864}0.76 & \cellcolor[HTML]{FD6864}0.87 & \cellcolor[HTML]{FD6864}0.77 \\ \cmidrule{3-12} 
                                     & \multirow{-3}{*}{Flat}   & V & \cellcolor[HTML]{FD6864}0.87 & \cellcolor[HTML]{32CB00}1.00 & \cellcolor[HTML]{FD6864}0.80 & \cellcolor[HTML]{FD6864}0.70 & \cellcolor[HTML]{FFCE93}0.91 & \cellcolor[HTML]{FD6864}0.69 & \cellcolor[HTML]{FD6864}0.30 & \cellcolor[HTML]{FD6864}0.75 & \cellcolor[HTML]{FD6864}0.30 \\ \cmidrule{2-12} 
                                     &                          & C & \cellcolor[HTML]{32CB00}0.99 & \cellcolor[HTML]{32CB00}1.00 & \cellcolor[HTML]{9AFF99}0.96 & \cellcolor[HTML]{FFCE93}0.95 & \cellcolor[HTML]{32CB00}0.98 & \cellcolor[HTML]{FD6864}0.86 & \cellcolor[HTML]{FD6864}0.66 & \cellcolor[HTML]{FFCE93}0.93 & \cellcolor[HTML]{FD6864}0.78 \\ \cmidrule{3-12} 
                                     &                          & F & \cellcolor[HTML]{32CB00}0.98 & \cellcolor[HTML]{32CB00}1.00 & \cellcolor[HTML]{32CB00}0.97 & \cellcolor[HTML]{FFCE93}0.94 & \cellcolor[HTML]{32CB00}0.99 & \cellcolor[HTML]{FD6864}0.52 & \cellcolor[HTML]{FD6864}0.76 & \cellcolor[HTML]{FD6864}0.82 & \cellcolor[HTML]{FD6864}0.77 \\ \cmidrule{3-12} 
                                     & \multirow{-3}{*}{Medium} & V & \cellcolor[HTML]{32CB00}0.98 & \cellcolor[HTML]{9AFF99}0.97 & \cellcolor[HTML]{9AFF99}0.95 & \cellcolor[HTML]{FD6864}0.88 & \cellcolor[HTML]{32CB00}1.00 & \cellcolor[HTML]{FD6864}0.68 & \cellcolor[HTML]{FD6864}0.32 & \cellcolor[HTML]{FD6864}0.75 & \cellcolor[HTML]{FD6864}0.33 \\ \cmidrule{2-12} 
                                     &                          & C & \cellcolor[HTML]{32CB00}1.00 & \cellcolor[HTML]{32CB00}1.00 & \cellcolor[HTML]{32CB00}0.99 & \cellcolor[HTML]{9AFF99}0.97 & \cellcolor[HTML]{32CB00}0.99 & \cellcolor[HTML]{FD6864}0.31 & \cellcolor[HTML]{FD6864}0.68 & \cellcolor[HTML]{FD6864}0.87 & \cellcolor[HTML]{FD6864}0.73 \\ \cmidrule{3-12} 
                                     &                          & F & \cellcolor[HTML]{32CB00}0.99 & \cellcolor[HTML]{32CB00}1.00 & \cellcolor[HTML]{32CB00}0.99 & \cellcolor[HTML]{FFCE93}0.94 & \cellcolor[HTML]{9AFF99}0.98 & \cellcolor[HTML]{FD6864}0.58 & \cellcolor[HTML]{FD6864}0.74 & \cellcolor[HTML]{FD6864}0.74 & \cellcolor[HTML]{FD6864}0.74 \\ \cmidrule{3-12} 
\multirow{-9}{*}{OW} & \multirow{-3}{*}{Rough}  & V & \cellcolor[HTML]{32CB00}1.00 & \cellcolor[HTML]{9AFF99}0.97 & \cellcolor[HTML]{32CB00}1.00 & \cellcolor[HTML]{FFCE93}0.92 & \cellcolor[HTML]{9AFF99}0.96 & \cellcolor[HTML]{FD6864}0.34 & \cellcolor[HTML]{FD6864}0.61 & \cellcolor[HTML]{FD6864}0.61 & \cellcolor[HTML]{FD6864}0.61 \\ \midrule
\end{tabular}
\vspace{1em} 
\caption{
\textbf{Normalized model scores across tasks and input modalities. }Models were evaluated on treadmill walking (TW), treadmill running (TR), and overground walking (OW) tasks with varying terrain. The input modalities (M) were: CoM-relevant kinematics (C), full-body kinematics (F), and visual gaze (V). The normalized model scores use the following color coding: dark green ($0.98$--$1.00$), light green ($0.95$--$0.98$), orange ($0.90$--$0.95$), and red ($< 0.9$). Normalized model scores, defined in the Methods section, are rounded to two decimal places, potentially resulting in two models having a score of $1.00$ (e.g., the other model's score could be $0.997$).
} 
\label{table:model-performance} 
\end{table}

%% file: discussion.tex
\section*{Discussion}
In this study, we discover that the timescale of locomotor control strategies depend on the environmental context as well as the sensory input modality. We present a data-driven framework to predict future motor actions from time-varying input modalities across real-world contexts. By employing deep learning architectures that incorporate the temporal history of state estimates, we outperform previous models in predicting future foot placements during non-stationary tasks. We compare the predictive power of various input modalities (e.g., center-of-mass kinematics, full-body kinematics, and gaze fixations) against a baseline model. Using this framework, we reveal a tradeoff between fast- and slow-timescale predictions across contexts, where humans rely more on fast timescale predictions in uncertain contexts such as rough terrain. We quantify the contribution of each modality to foot placement through their distinct prediction timescales, finding that gaze predicts future foot placements before the body states, and the full-body states predict foot placement before the center-of-mass states. Finally, we identify the swing phase at which the foot begins to carry significant predictive information about the future foot placement, which does not necessarily correspond to the start of the swing phase. The models we developed can be integrated with locomotor rehabilitation technologies and simulations to improve their generality across environmental contexts. This framework can be extended to other input modalities and motor actions to characterize real world motor behavior. 

\paragraph{Comparing different model architectures in their ability to capture locomotor control.} The model performances across tasks and input modalities (Figure \ref{table:model-performance}) highlight the strengths and limitations of different network architectures. During laboratory-constrained treadmill locomotion, linear models demonstrate performance comparable to the best-performing nonlinear models. This suggests that control strategies during treadmill locomotion are relatively simple and low dimensional \cite{kuvulmaz2005time, hyndman2018forecasting}, and therefore can be effectively captured by previous models that assume fixed-timescale linear mappings \cite{wang2014stepping, seethapathi2019step}. Interestingly, the Temporal Convolutional Network (TCN) performed worse than linear models, which can be attributed to its rigid structural design, or more specifically, its reliance on fixed-size convolutional kernels and local receptive fields, making it more prone to overfitting to noise. During treadmill locomotion tasks, where there is less input history-dependence, fully connected neural networks (FCNN) perform well as a result of the simplicity and linearity of the mappings \cite{nguyen2018interpretable}. During overground walking, we observe that GRUs \cite{cho2014learning} and Transformers \cite{vaswani2017attention} perform the best, since their gating or attention mechanisms enable the models to dynamically focus on the most relevant timesteps. This implies that when the task demands are more non-stationary i.e. when the statistics of the environment are time-varying, this will influence the timescale of the input history dependence for control. While previous models provide valuable insights into foot placement control in constrained environments like treadmill walking \cite{wang2014stepping, seethapathi2019step, joshi2019controller}, they fail to capture the complexity of more naturalistic locomotion such as overground walking. During overground walking, TCNs show comparable performance to other nonlinear models as their ability to process long-term dependencies becomes more relevant \cite{bai2018empirical, lea2017temporal}. In contrast, the FCNN now performs the worst among nonlinear models, highlighting the benefits of architectures that specialize in time series prediction in non-stationary scenarios. This is consistent with machine learning research in other domains which show that models explicitly designed for sequential data often outperform generic architectures in handling temporal dependencies \cite{franceschi2019unsupervised, ismail2019deep, bagnall2017great}. 

\paragraph{Tradeoffs between fast- and slow-timescale action predictions.} 
Our study identifies fast- and slow-timescale control strategies hidden in real-world movement data (Figure \ref{fig:fig3}A). Previous research has highlighted the importance of fast-timescale within-step corrections, such as those used for obstacle avoidance or unexpected terrain changes to enable stable locomotion \cite{daley2006running, van2007muscle, wang2014stepping}, showing that these corrections have a stabilizing effect on the body. Studies that have analyzed gaze fixations, on the other hand, have identified slower-timescale control strategies \cite{matthis2018gaze}, but have not tested whether these co-exist with or replace fast-timescale strategies. This gap is important to address, because the hierarchical motor control hypothesis posits the existence of both low-level fast timescale and high-level slow timescale processes \cite{merel2019hierarchical, seethapathi2024exploration}. Here, we discover the existence of both gaze-predictive slow timescale and body state-predictive fast timescale control strategies during natural locomotion. For walking on uneven terrain with varying complexity, we discover a context-dependent tradeoff between slow- and fast-timescale prediction (Figure \ref{fig:fig3}C). We interpret this tradeoff to reflect the greater prioritization of fast-timescale strategies in more complex environments, due to the rapid changes in the environmental statistics as a function of time. This tradeoff could be explained as the outcome of an optimization process in which working-memory resources used to represent the environment are allocated efficiently \cite{hayhoe2018control}, relying on more recent information in more complex environments. Indeed, while optimal feedback control suggests that there is more reliance on feedback in the presence of uncertainty \cite{kuo2002relative, todorov2002optimal, maurus2024increased}, here we provide evidence that could help extend this theory to multi-timescale control (Figure \ref{fig:fig3}C). Our model advances our understanding of how predictive planning and reactive corrections are combined in dynamic environments, characterizing how the sensorimotor control system integrates fast- and slow-timescale strategies to achieve adaptive movements. 

\paragraph{Input modality-dependent timescales for locomotor control.} By analyzing the relative predictive power of different input modalities our approach identifies when each of them becomes useful for controlling future actions (Figure \ref{fig:fig4}). Specifically, we find that full-body kinematics predict future foot placement earlier than CoM-relevant kinematics during treadmill running and overground walking. This earlier full-body prediction could be explained by the hypothesis that preparatory actions distributed across body segments may provide early cues to help modulate the CoM’s state \cite{nayeem2021preparing}. Alternatively, earlier full-body predictions could be explained by the correlation with features not captured by the CoM state such as whole-body angular momentum and its role in foot placement prediction \cite{leestma2023linking, van2025simultaneous}. We also found that the utilization of CoM-relevant kinematics is delayed during overground walking compared to treadmill walking. This delay likely stems from the increased variability in terrain and the demands of path planning in overground environments, which require more time to integrate information needed to determine optimal foot placement \cite{maclellan2006adaptations, hak2013steps}. Additionally, as terrain roughness increases, there is a greater reliance on gaze, as evidenced by the increasing relative predictive power of gaze fixations as an input modality \cite{matthis2017critical, patla1997understanding}. Previous work has demonstrated that humans use visual information for path planning on complex terrain, actively avoiding steep steps with large height changes in favor of flatter, more circuitous paths \cite{muller2024foothold}. Our findings further suggest that gaze information primarily predicts lateral (Figure \ref{fig:fig4}A) rather than fore-aft foot placement (Figure \ref{fig:fig8}) on uneven terrains, which aligns with previous findings that humans tend to adjust step width rather than step length to avoid obstacles \cite{schulz2012healthy, collins2013two}.

\paragraph{Conclusions and Future Work.} Our work provides a data-driven framework for understanding the control timescales at play during locomotion by combining real-world movement data with machine learning approaches. By predicting future actions from past input states, our framework characterizes these control timescales and demonstrates that nonlinear models consistently outperform traditional linear models across various environmental contexts. This approach is comprehensive and generalizable, allowing for a systematic comparison of how different input modalities and contexts influence control. By pinpointing the timescales at which past states best predict future actions, our work offers data-driven insights into the temporal structure of motor planning and control in complex, high-dimensional tasks. Our framework can be extended in future work to analyze large-scale datasets from different motor behaviors \cite{taheri2020grab, kratzer2020mogaze} and species \cite{schwaner2021future, muramatsu2025wildpose}. The context- and modality-dependent timescales identified by our approach can also be integrated into physics-based simulations of human movement. This integration could lead to more human-like predictive simulations across different environmental contexts and help test hypotheses in-silico about the sensory and neural basis of the prediction timescales we discover (Figure \ref{fig:expo}A) \cite{song2015neural, song2017evaluation, caggiano2022myosuite}. Additionally, these data-driven models can be integrated into rehabilitation technologies to improve their applicability to everyday settings. They can be integrated into real-time biofeedback systems \cite{giggins2013biofeedback, hribernik2022review} to provide predictive feedback about body states, helping to determine which states to relay and at what timescale. Similarly, our models can be combined with wearable robots , enabling them to simultaneously use both gaze \cite{bao2020vision} and body state \cite{kang2019real} signals to inform planned assistive profiles.

%% file: method.tex
\section*{Methods} 
In this section, we present our data-driven framework for analyzing and inferring control timescales across environmental contexts and input modalities. First, we describe the datasets used in this study and how the inputs and outputs for the models were structured. Next, we introduce the model architectures used to predict foot placement from the different input modalities. We further provide details on the training and evaluation procedures for these models, and discuss the statistical tests employed to examine the timescales. 

\subsection*{Data Description and processing}
Our framework is versatile and can be applied to any dataset with time-series inputs and discrete output events. To demonstrate this flexibility, we utilized several existing datasets of human locomotion across a range of contexts (Figure \ref{fig:fig1}A), including walking and running on both treadmills and overground, as well as across uneven terrain. The treadmill walking dataset was collected by Wang and Srinivasan \cite{wang2014stepping}, while the treadmill running dataset on even and uneven terrains was collected by Voloshina and colleagues \cite{voloshina2015biomechanics}. The overground walking dataset, collected by Matthis and colleagues \cite{matthis2018gaze}, includes full body kinematics of individuals walking on flat, medium, and rough terrains alongside gaze information recorded to track ground fixations during locomotion. 

For all studies, we processed kinematic data using a zero-lag 4th order Butterworth low-pass filter (cutoff frequency of 6 Hz). Here, let $x$ denote the lateral direction, $y$ denote the fore-aft direction, and $z$ denote the vertical direction. For treadmill movement datasets, we adjusted the fore-aft marker positions to account for belt speed to allow direct comparisons with overground data. Formally, let $y(t)$ denote the forward position of any marker at time $t$, the adjusted fore-aft position is defined as $y'(t) = y(t) + v \cdot t$, where $v$ is the belt speed. For overground data, the fore-aft axis was aligned with the main direction of locomotion, such that the pelvis position starts at $(0, 0)$ on the x-y plane and moved towards the positive y-axis. We computed the velocity of each marker using 4th-order centered finite differences of their positions. Next, we computed the heel-strike timings as the time at which the fore-aft distance between the foot and the pelvis markers is maximal \cite{banks2015using}. While other gait segmentation procedures based on velocity threshold produce similar estimates of contact detection \cite{zeni2008two}, we chose to adopt the method based on the relative foot position as it was more robust to noise during overground locomotion. We segmented the data into individual gait cycles and temporally interpolated them into 20 equally spaced gait phases. We implemented tests to detect and discard abnormal gait cycles. These tests verified whether the stance foot was on the ground during the stance phase (walking data only), and whether the timing of each heel-strike was reasonable relative to neighboring heel-strikes. Overall, we discarded fewer than 0.1\% of the steps for treadmill walking and running, and fewer than 2\% of the steps for the overground locomotion. 

\subsection*{Modeling framework}
Foot placement control can depend on a history of input modalities, such as postural or gaze information. In this study, we developed a data-driven framework to evaluate the ability of various input modalities to predict future foot placement (see Figure \ref{fig:fig1} for an overview of the modeling framework). 

We would like to analyze whether a history of inputs are integrated throughout locomotion to inform future foot placements. These inputs can include CoM-relevant kinematics, swing foot kinematics, full-body kinematics, and gaze fixations (Figure \ref{fig:fig1}B). The output predicted by the models consists of lateral and fore-aft foot placement relative to the contact location of the opposite foot at heel-strike (Figure \ref{fig:fig1}C). In this study, we leveraged data-driven models to evaluate the relationships between these input modalities and future foot placement. Let $v_i\in\{0, 1, \dots, T-1\}$ be the trial ID of data point $i$, where T is the number of trials, and $l_i \in \left\{ 0,1\right\}$ be the left-right (L/R) flag of data point $i$, representing whether the corresponding output data point is a left (0) or a right (1) heel-strike. The time series capturing the input modalities is denoted by $\mathbf{S}_i=\left[\mathbf{s}_i^{(1)}, \mathbf{s}_i^{(2)}, \dots, \mathbf{s}_i^{(m)}\right]$, where m is the number of features, and $\mathbf{s}_i^{(j)}$ is the $j$-th feature of the time series. This time series starts 6 steps prior to the predicted heel-strike (Figure \ref{fig:fig1}D) and its end is sampled from the previous gait cycle with 20 equally spaced gait phases. This range is chosen because 6 steps is a reasonable upper bound for the amount of information that can be used to plan future foot placement, while 2 steps is a reasonable lower bound  \cite{wang2014stepping, seethapathi2019step, matthis2018gaze}. Let $\phi$ denote the gait phase relative to one gait cycle prior to heel-strike (i.e. previous heel-strike of the same foot). With the dataset sampled at 20 timesteps per gait cycle, the length of $\mathbf{s}_i^{(j)}$ ranges from 41 (when time series ends at $\phi=0$) to 61 (when time series ends at $\phi=1$). We define the concatenated inputs as $X=\{x_i\}_{i=1}^n$, where $x_i=\{\mathbf{S}_i,v_i,l_i\}$, and seek to obtain the mapping between those inputs $X$ and the outputs $Y=\{y_i\}_{i=1}^n$, where 
$y_i=\left[f_i^{ML},f_i^{AP}\right]$ where $f_i^{ML}$ and $f_i^{AP}$ represent the mediolateral (ML) and anteroposterior (AP) foot placement, respectively.

\subsubsection*{Nonlinear models} \label{subsubsec:nonlinear} 
Training deep neural networks typically requires large amounts of data \cite{hestness2017deep}, which is challenging for human locomotion in everyday environments as most datasets only capture a few minutes of data for each trial. Therefore, it is often impractical to train deep learning models tailored to specific individuals or trials. To address this limitation, we pooled the data from all trials for a given environmental context and trained a single model that includes trial ID as an input. By doing so, the model can learn trial-specific variations in a data-driven way while efficiently leveraging the entire dataset (Figure \ref{fig:fig1}E). We first reshaped the left-right flag $l_i$ to match the time series input's dimension and concatenated them, which resulted in a matrix of shape $(t, m+1)$, where $m$ is the number of features in the time series. The trial ID was embedded \cite{guo2016entity} into a vector $E$ of length $\left|E\right|=\lceil \sqrt{T} \rceil$ and integrated with the time series data in a manner that is dependent on how each neural network processes the timesteps. The embedding enables us to isolate the inter-trial variability from the context- and modality-specific variability, which are the focuses of this study. The models we used include: 

\begin{enumerate}
    \item Long Short-Term Memory Model (LSTM) \cite{hochreiter1997long}: The trial embedding is passed through a fully connected layer to transform it into a vector matching the size of the LSTM hidden layer. This transformed embedding is used to initialize the hidden state of the LSTM \cite{karpathy2015deep}. The LSTM then processes the time series sequentially with gating mechanisms, leveraging this trial-specific initialization. Finally, the output of the LSTM is passed through a fully connected layer with a rectified linear unit (ReLU) to predict the foot placement. 
    \item Gated Recurrent Units (GRU) Model \cite{cho2014learning}: Similar to LSTM with the exception that the LSTM architecture is replaced by a GRU one. Specifically, GRU has fewer parameters compared to an LSTM model. 
    \item Temporal Convolutional Network (TCN) \cite{bai2018empirical}: Unlike the LSTM and GRU models, the TCN processes the time series using convolutional layers to capture temporal dependencies in parallel instead of sequentially. The trial embedding is concatenated with the time series along the feature axis before being passed into the TCN. The TCN architecture consists of two dilated convolutional layers with downsampling and residual connections to effectively model temporal dependencies. The output of the TCN block is then passed through fully connected layers with ReLU activation to predict foot placement.
    \item Transformer \cite{vaswani2017attention}: The Transformer model leverages self-attention mechanisms to capture temporal dependencies in the time series. The trial embedding is concatenated with the time series along the time axis, allowing the model to incorporate trial-specific information. The architecture includes a feature embedding layer, positional encoding to preserve temporal order, and a Transformer encoder consisting of multi-head self-attention and feedforward layers. The output of the encoder is aggregated and passed through two fully connected layers with ReLU activation to predict the foot placement. 
    \item Fully connected neural network (FCNN): We explored architectures not specifically designed for time series data. In this approach, the time series is first flattened into a single vector, resulting in a feature vector of shape $tk+\left|E\right|+1$, where $t$ is the number of timesteps, $k$ is the number of features, and $\left|E\right|$ is the dimension of the trial embedding. This vector is then passed through a fully connected neural network with multiple layers and ReLU activations. The network architecture employs a progressive reduction in the number of nodes per layer, controlled by a decay parameter, which determines both the total number of layers and the rate at which the node count decreases in each subsequent layer.
\end{enumerate}
\subsubsection*{Linear models}
For linear models, we train one model for each trial for two key reasons. First, linear models are less data-intensive and can effectively be trained on the limited data available within individual trials. Second, linear models lack the representational capacity to leverage trial-specific embeddings. By training trial-specific models, we ensure that the unique relationships within each trial are accurately captured and that the models are evaluated fairly.

\begin{enumerate}
    \item Linear instance (LI): Instead of analyzing the entire time series, we focus on a linear model that utilizes only the last instance of the time series (i.e. the body state at a particular gait phase). We fit a linear regression model to the last row vector of the time series input. This approach is similar to the models presented in \cite{wang2014stepping}. 
    \item Linear history (LH): The time series input is flattened into a vector of shape $tk$, and an Ordinary Least Squares regression is fitted to the data. 
    \item Linear instance with L2 regularization (LI2): L2 regularization (Ridge regression) is applied to the linear instance model to reduce overfitting. 
    \item Linear history with L2 regularization (LH2): L2 regularization (Ridge regression) is applied to the linear history model to reduce overfitting. 
\end{enumerate} 

\subsection*{Model evaluation and baseline model} \label{subsec:nest}
We employed nested cross-validation to evaluate the models' performance and hyperparameter tuning, ensuring an unbiased estimate of model performance and preventing data leakage \cite{bates2024cross}. We initially split the dataset into five equal folds; in each iteration of the outer loop, 80\% of the data was allocated for training and validation, while the remaining 20\% was held out as the testing set. The training/validation set was further subdivided into five inner folds. Four inner folds were used to train the model, and the remaining fold was used for early stopping, with a patience of 50 epochs. This process was repeated for every combination of hyperparameters, with parameter updates performed using the ADAM optimizer \cite{kingma2014adam} and a learning rate of 0.001. We used mean squared error (MSE) as the loss function. After each training session, the model's performance was evaluated on the outer testing fold. The outer loop was repeated five times, ensuring that each data point in the original dataset was used once for testing and four times for training/validation.

We first identified optimal hyperparameters using grid search (or random search for large hyperparameter spaces exceeding 100 configurations; see Appendix for details). These hyperparameters were then used in the nested cross-validation process. Aggregated predictions from all outer folds, covering 100\% of the data points, were used to compute the explained variance ($R^2$) of foot placement predicted by the input modality. Under this nested cross-validation framework, it is possible, though unlikely, to obtain negative $R^2$ values, which occurs when the model fails to generalize to the testing dataset, performing worse than simply predicting the average foot placements (which would result in $R^2=0$). In such cases, the model's predictive power is minimal or nonexistent. To prevent overfitting, we used early stopping and dropout when training the neural networks, whereas $L2$ regularization is used to prevent overfitting to historical data or high-dimensional input space for linear models. Since neural networks are more complex than linear models, they inherently exhibit lower bias and higher variance. To mitigate this higher variance of neural networks and to better interpret model behavior across gait phases, we smooth the $R^2$ curves using local regression (LOWESS\cite{cleveland1979robust}) and cubic spline. 

To evaluate model performance, we assigned a score to each model based on the Root Mean Squared Error (RMSE). As the time window increases, the input contains accumulative information about the foot placement thereby enhancing the models' predictive power. We used the RMSE curve achieved by the swing foot kinematics as a baseline, reflecting the predictive information for future foot placement inherently contained in the swing foot dynamics \cite{takens2006detecting}. We define the model score as the RMSE as defined in the next section.  The normalized model score is then computed relative to the model with the highest score. Formally, let $\text{RMSE}(m, \phi)$ denote the RMSE of model $m$ at gait phase $\phi$ and let $c$ denote the gait phase at the end of the input window. We define the model score $s_m$:
$$s_m=\frac{1}{\left|\psi\right|}\sum_{\phi\in\psi}\frac{\min_{m'}\text{RMSE}(m',\phi)}{\text{RMSE}(m,\phi)}$$ 
\noindent where $\psi = \{0, 0.05, 0.10, \dots, c\}$, and the normalized model score $s_m^n$: 
$$s_m^n=\frac{s_m}{\max_{m'}s_{m'}}$$
Note that the normalized model score is bounded within the interval $(0, 1]$, where the lower bound is approached when the RMSE is arbitrarily large, and the upper bound is achieved by the optimal model. 

\subsection*{Quantifying the prediction timescale} \label{subsec:control-timescale} 
\begin{figure}[htp] 
    \centering
    \includegraphics[width=0.68\linewidth]{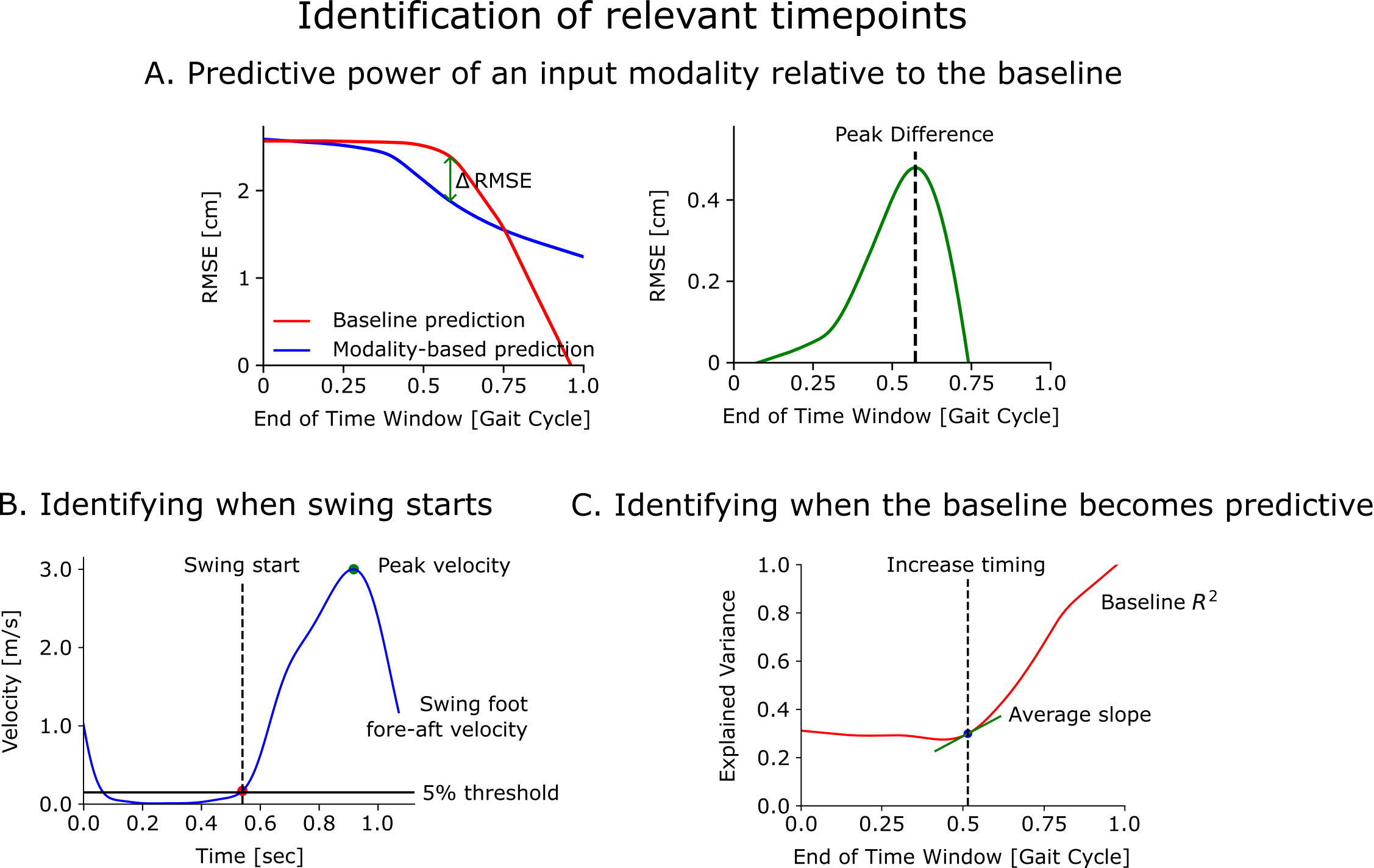}
    \caption{\textbf{Identification of relevant timepoints.} \textbf{A} The relative RMSE between modality-based predictions and the autoregressive baseline prediction as a function of gait phases. The timing of the peak difference indicates the phase where the input modality is the strongest predictor. \textbf{B} An example velocity curve along the locomotor axis, with the 5\% of peak velocity threshold used to determine the onset of the swing phase. \textbf{C} The baseline $R^2$ curve, showing a breakpoint where $R^2$ increases significantly. The timing of this breakpoint is identified as the phase where the slope exceeds the average slope. }
    \label{fig:fig2}
\end{figure}

The predictive power of foot placement using different input modalities across the gait cycle provides insight into the timescale of the control strategy. For an input modality to be considered a strong predictor, it must outperform the autoregressive baseline and demonstrate an earlier and larger increase in predictive power. To evaluate this, we identify the gait phase where the input modality achieves its maximum performance over the baseline, measured by the relative RMSE gap (Figure \ref{fig:fig2}A). Beyond this phase, the input modality's contribution diminishes, as the predictive power of the swing foot itself increases at a faster rate. 

To better understand swing foot prediction timescale, we aim to identify two critical timing events: (1) the time at which the swing foot begins its swing phase and (2) the time at which it starts containing significant information about future foot placement. The onset of the swing phase is defined as the peak velocity of the swing foot along the fore-aft direction. From this peak, we trace backward to identify the point where the velocity drops below 5\% of its maximal value (Figure \ref{fig:fig2}B). This approach of anchoring the swing detection to the peak velocity ensures robustness and allows to avoid false positives. 

We analyze the autoregressive baseline $R^2$ curve to determine the time at which the swing foot kinematics begins to substantially predict future foot placement. The $R^2$ curve typically shows a non-decreasing trend as the gait cycle progresses, eventually reaching $R^2=1$ at the final timestep. A notable ``breakpoint'' in the $R^2$ curve is the phase at which the predictive power of the swing foot kinematics sharply rises. To systematically identify this breakpoint, we calculate the average slope of the $R^2$ curve, defined as $r = R^2(1) - R^2(0)$, and locate the gait phase where the slope reaches its maximal value. We then trace backward from this gait phase to find the first point where the slope falls below the average slope, marking the critical transition phase (Figure \ref{fig:fig2}C). This approach provides a robust method for identifying key transitions in the gait cycle and aligns with the underlying dynamics of gait and motor control.

%% file: acknowledgements.tex
\section*{Acknowledgments} 
Wei-Chen Wang was supported by a Mathworks EECS Fellowship and MIT Research Support Committee Grant. Antoine De Comite was supported by a K. Lisa Yang Ingergrative Computational Neuroscience (ICoN) Fellowship. This work is also supported by the McGovern Institute for Brain Research and the NSF Center for Integrative Movement Sciences Summer Institute. The authors declare no competing interests. 

%% file: appendices.tex
\section*{Supplementary Materials} 
\paragraph{Hyperparameter Tuning}
\label{app:hyperparameter-tuning} 
To ensure a fair comparison, we independently tune the hyperparameters for each model and input modality across all contexts using nested cross-validation \cite{yang2020hyperparameter, cawley2010over}. For hyperparameter optimization, we use grid search when the search space is manageable, and random search \cite{bergstra2012random} when the hyperparameter space exceeds 100 configurations. Below, we outline the hyperparameter search space for each model. See the Methods section for details on model architecture. 
\begin{itemize}
    \item LSTM: the size of the LSTM hidden layer, \texttt{hidden\_dim} $\in\{2, 4, 8, 16, 32, 64, 128, 256\}$. 
    \item GRU: the size of the GRU hidden layer, \texttt{hidden\_dim} $\in\{2, 4, 8, 16, 32, 64, 128, 256\}$. 
    \item FCNN: the fully connected layers are designed such that the number of nodes in each layer decays exponentially, with a minimum of 8 nodes per layer and a guaranteed minimum number of layers. The decay factor, \texttt{decay}$\in\{2, 4, 8, 16\}$; the dropout rate, \texttt{dropout}$\in\{0, 0.1, 0.2, 0.3\}$. 
    \item TCN: the size of the TCN block output size, \texttt{hidden\_dim} $\in\{4, 8, 16\}$; the kernel size, \texttt{kernel} $\in\{1, 3, 5, 7\}$; dilation factor, \texttt{dilation} $\in\{1, 2, 4\}$; the dropout rate \texttt{dropout} $\in\{0, 0.1, 0.2, 0.3\}$. 
    \item Transformer: the size of positional encoding, \texttt{hidden\_dim} $\in\{16, 32, 64\}$; the number of layers, \texttt{num\_layers} $\in\{2, 3, 4\}$; the number of attention heads, \texttt{num\_heads} $\in\{2, 4, 8\}$; the size of feedforward dimension, \texttt{ff\_dim} $\in\{16, 32, 64\}$; the dropout rate \texttt{dropout} $\in\{0, 0.1, 0.2, 0.3\}$. 
\end{itemize}
After determining the optimal hyperparameters for each context, input modality, and model, we evaluate the predictive power with the identified hyperparameter configuration. 
\paragraph{Predictive power across models} 
The Methods section describes how the performances of the models are evaluated. 

\begin{figure}[htp] 
    \centering
    \includegraphics[width=0.99\linewidth]{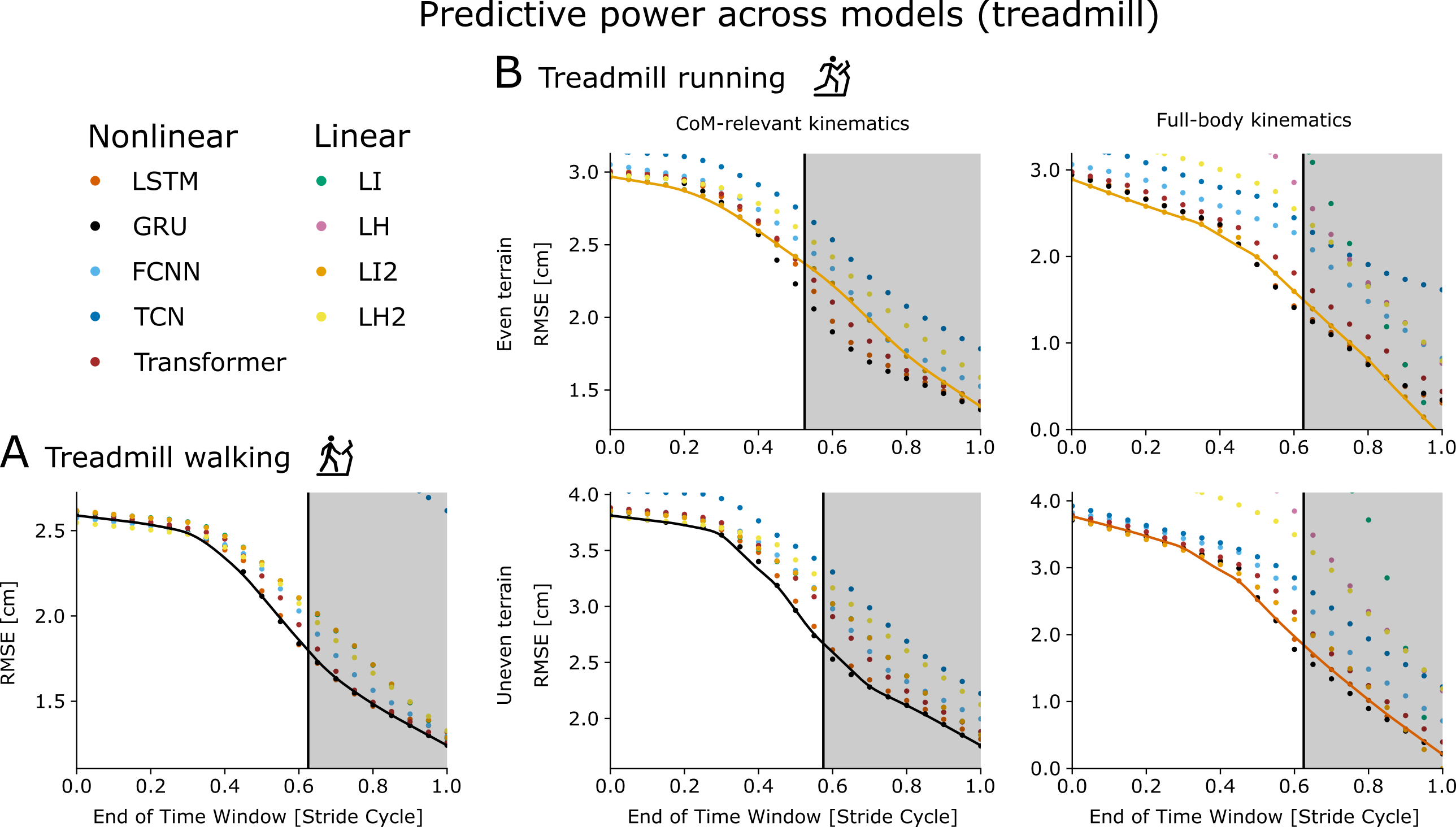}
    \caption{\textbf{Predictive power across models. }
    \textbf{A} Predictive power during treadmill walking. The vertical line indicates the critical phase when the relative predictive power is maximized, thereby only gait phases before the gray-shaded area are considered. The model with the average highest score prior to the critical phase was deemed optimal, and its predictive power is shown by the interpolated curve (LOWESS\cite{cleveland1979robust}). See the Methods section for more details on the evaluation metric. The normalized model score is summarized in Table \ref{table:model-performance}. 
    \textbf{B} Predictive power during treadmill running. }
    \label{fig:fig6}
\end{figure} 
\begin{figure}[htp] 
    \centering
    \includegraphics[width=0.97\linewidth]{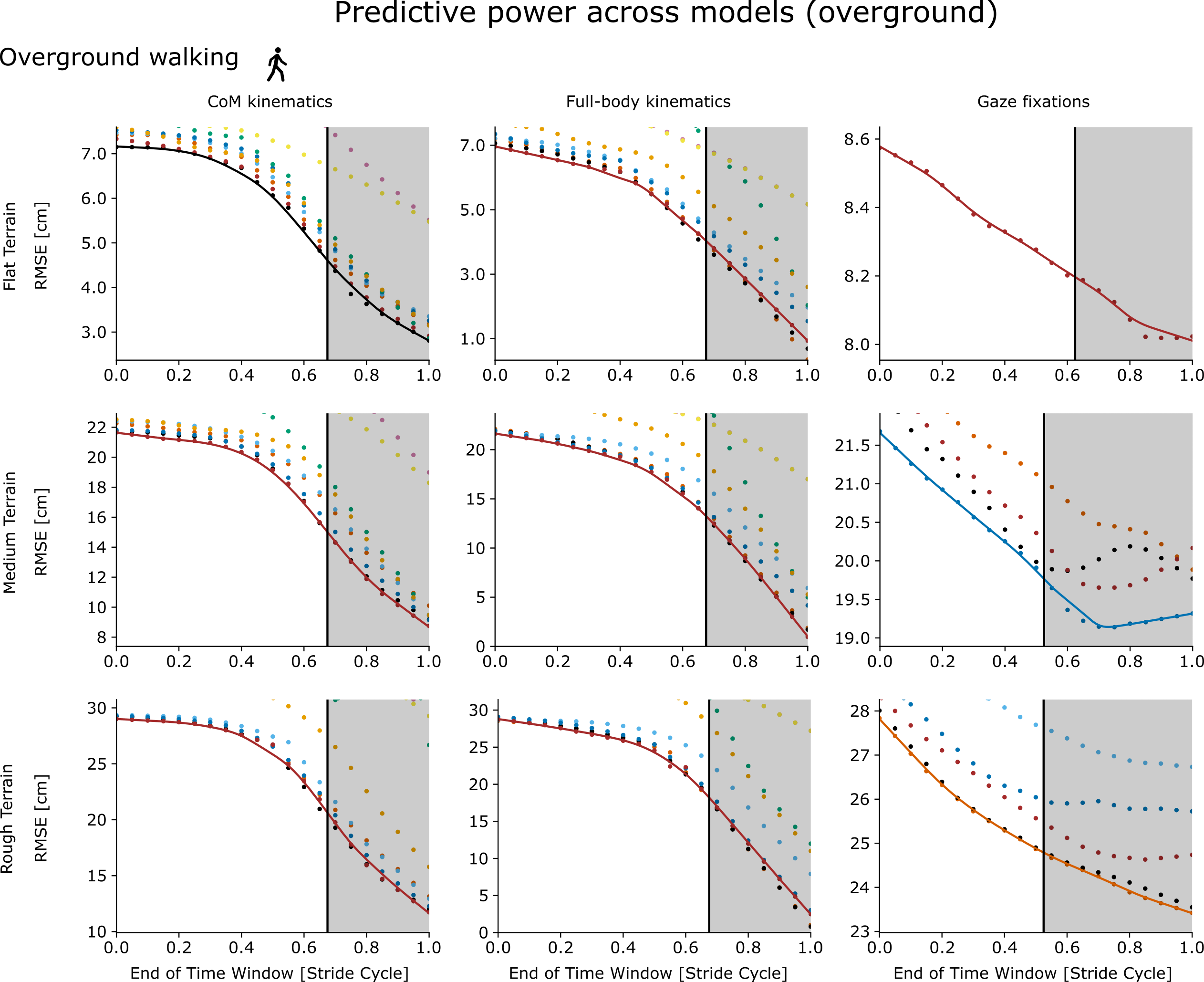}
    \caption{\textbf{Predictive power across models during overground walking.} See Figure \ref{fig:fig6} for details.}
    \label{fig:fig7}
\end{figure} 
\begin{figure}[htp] 
    \centering
    \includegraphics[width=0.85\linewidth]{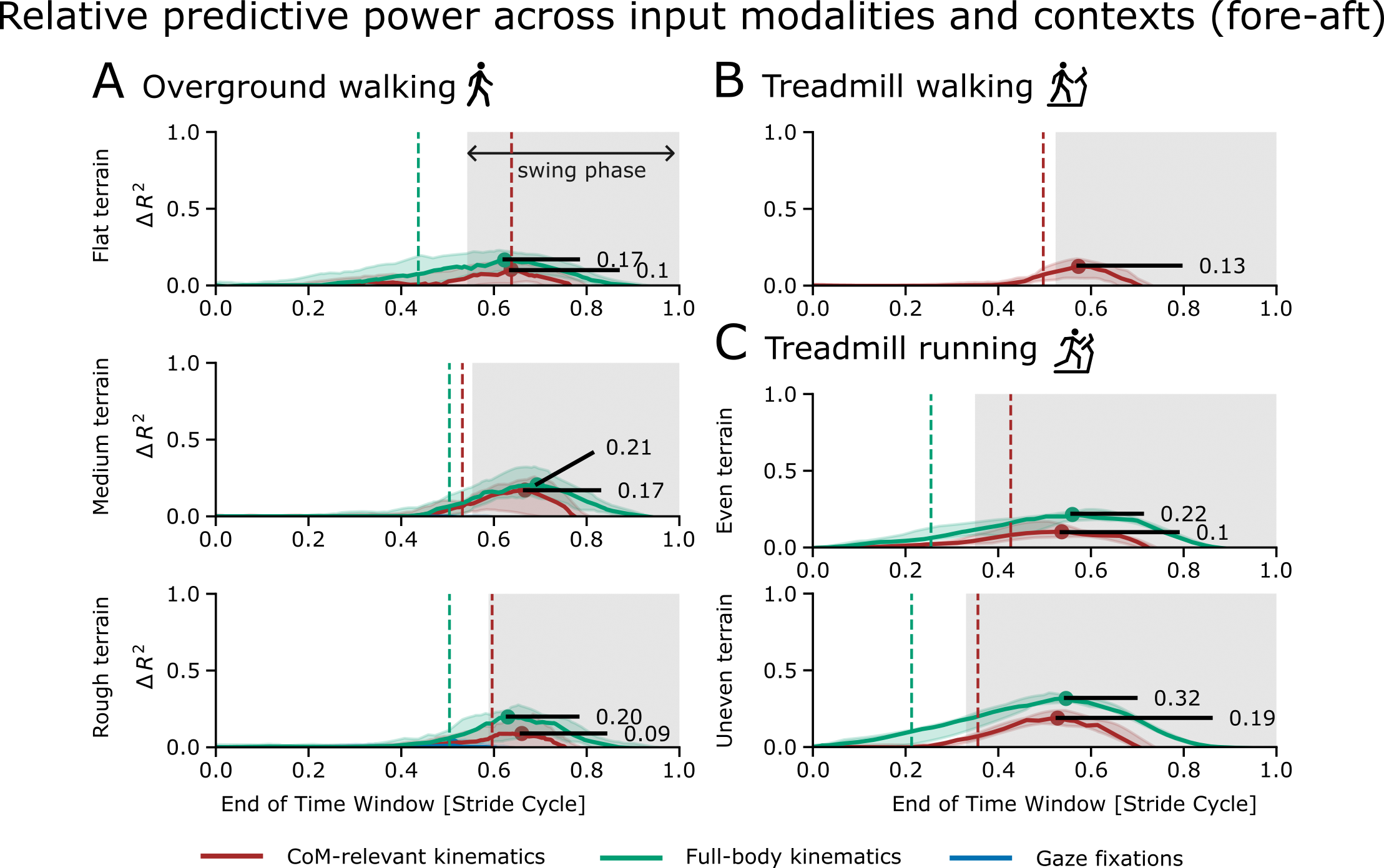}
    \caption{\textbf{Relative predictive power across input modalities and contexts} for fore-aft foot placement predictions during \textbf{A} Treadmill walking, \textbf{B} treadmill running, and \textbf{C} Overground walking. See Figure \ref{fig:fig3} for details. }
    \label{fig:fig8}
\end{figure}

%% file: main.bbl
\begin{thebibliography}{10}
\urlstyle{rm}
\expandafter\ifx\csname url\endcsname\relax
  \def\url#1{\texttt{#1}}\fi
\expandafter\ifx\csname urlprefix\endcsname\relax\def\urlprefix{URL }\fi
\expandafter\ifx\csname doiprefix\endcsname\relax\def\doiprefix{DOI: }\fi
\providecommand{\bibinfo}[2]{#2}
\providecommand{\eprint}[2][]{\url{#2}}

\bibitem{townsend1985biped}
\bibinfo{author}{Townsend, M.~A.}
\newblock \bibinfo{journal}{\bibinfo{title}{Biped gait stabilization via foot placement}}.
\newblock {\emph{\JournalTitle{Journal of biomechanics}}} \textbf{\bibinfo{volume}{18}}, \bibinfo{pages}{21--38} (\bibinfo{year}{1985}).

\bibitem{bruijn2018control}
\bibinfo{author}{Bruijn, S.~M.} \& \bibinfo{author}{Van~Die{\"e}n, J.~H.}
\newblock \bibinfo{journal}{\bibinfo{title}{Control of human gait stability through foot placement}}.
\newblock {\emph{\JournalTitle{Journal of The Royal Society Interface}}} \textbf{\bibinfo{volume}{15}}, \bibinfo{pages}{20170816} (\bibinfo{year}{2018}).

\bibitem{rankin2014neuromechanical}
\bibinfo{author}{Rankin, B.~L.}, \bibinfo{author}{Buffo, S.~K.} \& \bibinfo{author}{Dean, J.~C.}
\newblock \bibinfo{journal}{\bibinfo{title}{A neuromechanical strategy for mediolateral foot placement in walking humans}}.
\newblock {\emph{\JournalTitle{Journal of neurophysiology}}} \textbf{\bibinfo{volume}{112}}, \bibinfo{pages}{374--383} (\bibinfo{year}{2014}).

\bibitem{horslen2014modulation}
\bibinfo{author}{Horslen, B.~C.}, \bibinfo{author}{Dakin, C.~J.}, \bibinfo{author}{Inglis, J.~T.}, \bibinfo{author}{Blouin, J.-S.} \& \bibinfo{author}{Carpenter, M.~G.}
\newblock \bibinfo{journal}{\bibinfo{title}{Modulation of human vestibular reflexes with increased postural threat}}.
\newblock {\emph{\JournalTitle{The Journal of physiology}}} \textbf{\bibinfo{volume}{592}}, \bibinfo{pages}{3671--3685} (\bibinfo{year}{2014}).

\bibitem{matthis2018gaze}
\bibinfo{author}{Matthis, J.~S.}, \bibinfo{author}{Yates, J.~L.} \& \bibinfo{author}{Hayhoe, M.~M.}
\newblock \bibinfo{journal}{\bibinfo{title}{Gaze and the control of foot placement when walking in natural terrain}}.
\newblock {\emph{\JournalTitle{Current Biology}}} \textbf{\bibinfo{volume}{28}}, \bibinfo{pages}{1224--1233} (\bibinfo{year}{2018}).

\bibitem{muller2024foothold}
\bibinfo{author}{Muller, K.~S.} \emph{et~al.}
\newblock \bibinfo{journal}{\bibinfo{title}{Foothold selection during locomotion in uneven terrain: Results from the integration of eye tracking, motion capture, and photogrammetry}}.
\newblock {\emph{\JournalTitle{eLife}}} \textbf{\bibinfo{volume}{12}} (\bibinfo{year}{2024}).

\bibitem{shadmehr2008computational}
\bibinfo{author}{Shadmehr, R.} \& \bibinfo{author}{Krakauer, J.~W.}
\newblock \bibinfo{journal}{\bibinfo{title}{A computational neuroanatomy for motor control}}.
\newblock {\emph{\JournalTitle{Experimental brain research}}} \textbf{\bibinfo{volume}{185}}, \bibinfo{pages}{359--381} (\bibinfo{year}{2008}).

\bibitem{richer2024mobile}
\bibinfo{author}{Richer, N.}, \bibinfo{author}{Bradford, J.~C.} \& \bibinfo{author}{Ferris, D.~P.}
\newblock \bibinfo{journal}{\bibinfo{title}{Mobile neuroimaging: What we have learned about the neural control of human walking, with an emphasis on eeg-based research}}.
\newblock {\emph{\JournalTitle{Neuroscience \& Biobehavioral Reviews}}} \textbf{\bibinfo{volume}{162}}, \bibinfo{pages}{105718} (\bibinfo{year}{2024}).

\bibitem{darici2023humans}
\bibinfo{author}{Darici, O.} \& \bibinfo{author}{Kuo, A.~D.}
\newblock \bibinfo{journal}{\bibinfo{title}{Humans plan for the near future to walk economically on uneven terrain}}.
\newblock {\emph{\JournalTitle{Proceedings of the National Academy of Sciences}}} \textbf{\bibinfo{volume}{120}}, \bibinfo{pages}{e2211405120} (\bibinfo{year}{2023}).

\bibitem{seethapathi2024exploration}
\bibinfo{author}{Seethapathi, N.}, \bibinfo{author}{Clark, B.~C.} \& \bibinfo{author}{Srinivasan, M.}
\newblock \bibinfo{journal}{\bibinfo{title}{Exploration-based learning of a stabilizing controller predicts locomotor adaptation}}.
\newblock {\emph{\JournalTitle{Nature Communications}}} \textbf{\bibinfo{volume}{15}}, \bibinfo{pages}{9498} (\bibinfo{year}{2024}).

\bibitem{wang2014stepping}
\bibinfo{author}{Wang, Y.} \& \bibinfo{author}{Srinivasan, M.}
\newblock \bibinfo{journal}{\bibinfo{title}{Stepping in the direction of the fall: the next foot placement can be predicted from current upper body state in steady-state walking}}.
\newblock {\emph{\JournalTitle{Biology letters}}} \textbf{\bibinfo{volume}{10}}, \bibinfo{pages}{20140405} (\bibinfo{year}{2014}).

\bibitem{seethapathi2019step}
\bibinfo{author}{Seethapathi, N.} \& \bibinfo{author}{Srinivasan, M.}
\newblock \bibinfo{journal}{\bibinfo{title}{Step-to-step variations in human running reveal how humans run without falling}}.
\newblock {\emph{\JournalTitle{Elife}}} \textbf{\bibinfo{volume}{8}}, \bibinfo{pages}{e38371} (\bibinfo{year}{2019}).

\bibitem{afschrift2021similar}
\bibinfo{author}{Afschrift, M.}, \bibinfo{author}{De~Groote, F.} \& \bibinfo{author}{Jonkers, I.}
\newblock \bibinfo{journal}{\bibinfo{title}{Similar sensorimotor transformations control balance during standing and walking}}.
\newblock {\emph{\JournalTitle{PLoS computational biology}}} \textbf{\bibinfo{volume}{17}}, \bibinfo{pages}{e1008369} (\bibinfo{year}{2021}).

\bibitem{radosavovic2025humanoid}
\bibinfo{author}{Radosavovic, I.} \emph{et~al.}
\newblock \bibinfo{journal}{\bibinfo{title}{Humanoid locomotion as next token prediction}}.
\newblock {\emph{\JournalTitle{Advances in Neural Information Processing Systems}}} \textbf{\bibinfo{volume}{37}}, \bibinfo{pages}{79307--79324} (\bibinfo{year}{2025}).

\bibitem{xiong2023probability}
\bibinfo{author}{Xiong, J.} \emph{et~al.}
\newblock \bibinfo{journal}{\bibinfo{title}{A probability fusion approach for foot placement prediction in complex terrains}}.
\newblock {\emph{\JournalTitle{IEEE Transactions on Neural Systems and Rehabilitation Engineering}}} \textbf{\bibinfo{volume}{31}}, \bibinfo{pages}{4591--4600} (\bibinfo{year}{2023}).

\bibitem{chen2021probability}
\bibinfo{author}{Chen, X.}, \bibinfo{author}{Zhang, K.}, \bibinfo{author}{Liu, H.}, \bibinfo{author}{Leng, Y.} \& \bibinfo{author}{Fu, C.}
\newblock \bibinfo{journal}{\bibinfo{title}{A probability distribution model-based approach for foot placement prediction in the early swing phase with a wearable imu sensor}}.
\newblock {\emph{\JournalTitle{IEEE Transactions on Neural Systems and Rehabilitation Engineering}}} \textbf{\bibinfo{volume}{29}}, \bibinfo{pages}{2595--2604} (\bibinfo{year}{2021}).

\bibitem{lee2023deep}
\bibinfo{author}{Lee, S.-W.} \& \bibinfo{author}{Asbeck, A.}
\newblock \bibinfo{journal}{\bibinfo{title}{A deep learning-based approach for foot placement prediction}}.
\newblock {\emph{\JournalTitle{IEEE Robotics and Automation Letters}}} \textbf{\bibinfo{volume}{8}}, \bibinfo{pages}{4959--4966} (\bibinfo{year}{2023}).

\bibitem{asogwa2022using}
\bibinfo{author}{Asogwa, C.~O.}, \bibinfo{author}{Nagano, H.}, \bibinfo{author}{Wang, K.} \& \bibinfo{author}{Begg, R.}
\newblock \bibinfo{journal}{\bibinfo{title}{Using deep learning to predict minimum foot--ground clearance event from toe-off kinematics}}.
\newblock {\emph{\JournalTitle{Sensors}}} \textbf{\bibinfo{volume}{22}}, \bibinfo{pages}{6960} (\bibinfo{year}{2022}).

\bibitem{wolpert2000computational}
\bibinfo{author}{Wolpert, D.~M.} \& \bibinfo{author}{Ghahramani, Z.}
\newblock \bibinfo{journal}{\bibinfo{title}{Computational principles of movement neuroscience}}.
\newblock {\emph{\JournalTitle{Nature neuroscience}}} \textbf{\bibinfo{volume}{3}}, \bibinfo{pages}{1212--1217} (\bibinfo{year}{2000}).

\bibitem{van1999integration}
\bibinfo{author}{Van~Beers, R.~J.}, \bibinfo{author}{Sittig, A.~C.} \& \bibinfo{author}{Gon, J. J. D. v.~d.}
\newblock \bibinfo{journal}{\bibinfo{title}{Integration of proprioceptive and visual position-information: An experimentally supported model}}.
\newblock {\emph{\JournalTitle{Journal of neurophysiology}}} \textbf{\bibinfo{volume}{81}}, \bibinfo{pages}{1355--1364} (\bibinfo{year}{1999}).

\bibitem{matthis2014visual}
\bibinfo{author}{Matthis, J.~S.} \& \bibinfo{author}{Fajen, B.~R.}
\newblock \bibinfo{journal}{\bibinfo{title}{Visual control of foot placement when walking over complex terrain.}}
\newblock {\emph{\JournalTitle{Journal of experimental psychology: human perception and performance}}} \textbf{\bibinfo{volume}{40}}, \bibinfo{pages}{106} (\bibinfo{year}{2014}).

\bibitem{dietz2002proprioception}
\bibinfo{author}{Dietz, V.}
\newblock \bibinfo{journal}{\bibinfo{title}{Proprioception and locomotor disorders}}.
\newblock {\emph{\JournalTitle{Nature Reviews Neuroscience}}} \textbf{\bibinfo{volume}{3}}, \bibinfo{pages}{781--790} (\bibinfo{year}{2002}).

\bibitem{roden2015hip}
\bibinfo{author}{Roden-Reynolds, D.~C.}, \bibinfo{author}{Walker, M.~H.}, \bibinfo{author}{Wasserman, C.~R.} \& \bibinfo{author}{Dean, J.~C.}
\newblock \bibinfo{journal}{\bibinfo{title}{Hip proprioceptive feedback influences the control of mediolateral stability during human walking}}.
\newblock {\emph{\JournalTitle{Journal of neurophysiology}}} \textbf{\bibinfo{volume}{114}}, \bibinfo{pages}{2220--2229} (\bibinfo{year}{2015}).

\bibitem{vetter2000context}
\bibinfo{author}{Vetter, P.} \& \bibinfo{author}{Wolpert, D.~M.}
\newblock \bibinfo{journal}{\bibinfo{title}{Context estimation for sensorimotor control}}.
\newblock {\emph{\JournalTitle{Journal of Neurophysiology}}} \textbf{\bibinfo{volume}{84}}, \bibinfo{pages}{1026--1034} (\bibinfo{year}{2000}).

\bibitem{hayhoe2018control}
\bibinfo{author}{Hayhoe, M.~M.} \& \bibinfo{author}{Matthis, J.~S.}
\newblock \bibinfo{journal}{\bibinfo{title}{Control of gaze in natural environments: effects of rewards and costs, uncertainty and memory in target selection}}.
\newblock {\emph{\JournalTitle{Interface focus}}} \textbf{\bibinfo{volume}{8}}, \bibinfo{pages}{20180009} (\bibinfo{year}{2018}).

\bibitem{patla1997understanding}
\bibinfo{author}{Patla, A.~E.}
\newblock \bibinfo{journal}{\bibinfo{title}{Understanding the roles of vision in the control of human locomotion}}.
\newblock {\emph{\JournalTitle{Gait \& posture}}} \textbf{\bibinfo{volume}{5}}, \bibinfo{pages}{54--69} (\bibinfo{year}{1997}).

\bibitem{matthis2017critical}
\bibinfo{author}{Matthis, J.~S.}, \bibinfo{author}{Barton, S.~L.} \& \bibinfo{author}{Fajen, B.~R.}
\newblock \bibinfo{journal}{\bibinfo{title}{The critical phase for visual control of human walking over complex terrain}}.
\newblock {\emph{\JournalTitle{Proceedings of the National Academy of Sciences}}} \textbf{\bibinfo{volume}{114}}, \bibinfo{pages}{E6720--E6729} (\bibinfo{year}{2017}).

\bibitem{warren2006dynamics}
\bibinfo{author}{Warren, W.~H.}
\newblock \bibinfo{journal}{\bibinfo{title}{The dynamics of perception and action.}}
\newblock {\emph{\JournalTitle{Psychological review}}} \textbf{\bibinfo{volume}{113}}, \bibinfo{pages}{358} (\bibinfo{year}{2006}).

\bibitem{bruijn2013assessing}
\bibinfo{author}{Bruijn, S.~M.}, \bibinfo{author}{Meijer, O.}, \bibinfo{author}{Beek, P.} \& \bibinfo{author}{van Dieen, J.~H.}
\newblock \bibinfo{journal}{\bibinfo{title}{Assessing the stability of human locomotion: a review of current measures}}.
\newblock {\emph{\JournalTitle{Journal of the Royal Society Interface}}} \textbf{\bibinfo{volume}{10}}, \bibinfo{pages}{20120999} (\bibinfo{year}{2013}).

\bibitem{joshi2019controller}
\bibinfo{author}{Joshi, V.} \& \bibinfo{author}{Srinivasan, M.}
\newblock \bibinfo{journal}{\bibinfo{title}{A controller for walking derived from how humans recover from perturbations}}.
\newblock {\emph{\JournalTitle{Journal of The Royal Society Interface}}} \textbf{\bibinfo{volume}{16}}, \bibinfo{pages}{20190027} (\bibinfo{year}{2019}).

\bibitem{hasaneini2014swing}
\bibinfo{author}{Hasaneini, S.~J.}, \bibinfo{author}{Macnab, C.~J.}, \bibinfo{author}{Bertram, J.~E.} \& \bibinfo{author}{Leung, H.}
\newblock \bibinfo{title}{Swing-leg retraction efficiency in bipedal walking}.
\newblock In \emph{\bibinfo{booktitle}{2014 IEEE/RSJ International Conference on Intelligent Robots and Systems}}, \bibinfo{pages}{2515--2522} (\bibinfo{organization}{IEEE}, \bibinfo{year}{2014}).

\bibitem{hasaneini2013optimal}
\bibinfo{author}{Hasaneini, S.~J.}, \bibinfo{author}{Macnab, C.~J.}, \bibinfo{author}{Bertram, J.~E.} \& \bibinfo{author}{Leung, H.}
\newblock \bibinfo{title}{Optimal relative timing of stance push-off and swing leg retraction}.
\newblock In \emph{\bibinfo{booktitle}{2013 IEEE/RSJ International Conference on Intelligent Robots and Systems}}, \bibinfo{pages}{3616--3623} (\bibinfo{organization}{IEEE}, \bibinfo{year}{2013}).

\bibitem{mcgeer1990passive}
\bibinfo{author}{McGeer, T.}
\newblock \bibinfo{journal}{\bibinfo{title}{Passive dynamic walking}}.
\newblock {\emph{\JournalTitle{The international journal of robotics research}}} \textbf{\bibinfo{volume}{9}}, \bibinfo{pages}{62--82} (\bibinfo{year}{1990}).

\bibitem{collins2008dynamic}
\bibinfo{author}{Collins, S.~H.}
\newblock \emph{\bibinfo{title}{Dynamic walking principles applied to human gait}}.
\newblock Ph.D. thesis, \bibinfo{school}{University of Michigan} (\bibinfo{year}{2008}).

\bibitem{voloshina2015biomechanics}
\bibinfo{author}{Voloshina, A.~S.} \& \bibinfo{author}{Ferris, D.~P.}
\newblock \bibinfo{journal}{\bibinfo{title}{Biomechanics and energetics of running on uneven terrain}}.
\newblock {\emph{\JournalTitle{The journal of experimental biology}}} \textbf{\bibinfo{volume}{218}}, \bibinfo{pages}{711--719} (\bibinfo{year}{2015}).

\bibitem{kuvulmaz2005time}
\bibinfo{author}{Kuvulmaz, J.}, \bibinfo{author}{Usanmaz, S.} \& \bibinfo{author}{Engin, S.~N.}
\newblock \bibinfo{title}{Time-series forecasting by means of linear and nonlinear models}.
\newblock In \emph{\bibinfo{booktitle}{MICAI 2005: Advances in Artificial Intelligence: 4th Mexican International Conference on Artificial Intelligence, Monterrey, Mexico, November 14-18, 2005. Proceedings 4}}, \bibinfo{pages}{504--513} (\bibinfo{organization}{Springer}, \bibinfo{year}{2005}).

\bibitem{hyndman2018forecasting}
\bibinfo{author}{Hyndman, R.}
\newblock \emph{\bibinfo{title}{Forecasting: principles and practice}} (\bibinfo{publisher}{OTexts}, \bibinfo{address}{Melbourne, Australia}, \bibinfo{year}{2018}).

\bibitem{nguyen2018interpretable}
\bibinfo{author}{Nguyen, T.~L.}, \bibinfo{author}{Gsponer, S.}, \bibinfo{author}{Ilie, I.} \& \bibinfo{author}{Ifrim, G.}
\newblock \bibinfo{journal}{\bibinfo{title}{Interpretable time series classification using all-subsequence learning and symbolic representations in time and frequency domains}}.
\newblock {\emph{\JournalTitle{arXiv preprint arXiv:1808.04022}}}  (\bibinfo{year}{2018}).

\bibitem{cho2014learning}
\bibinfo{author}{Cho, K.} \emph{et~al.}
\newblock \bibinfo{journal}{\bibinfo{title}{Learning phrase representations using rnn encoder-decoder for statistical machine translation}}.
\newblock {\emph{\JournalTitle{arXiv preprint arXiv:1406.1078}}}  (\bibinfo{year}{2014}).

\bibitem{vaswani2017attention}
\bibinfo{author}{Vaswani, A.} \emph{et~al.}
\newblock \bibinfo{journal}{\bibinfo{title}{Attention is all you need}}.
\newblock {\emph{\JournalTitle{Advances in neural information processing systems}}} \textbf{\bibinfo{volume}{30}} (\bibinfo{year}{2017}).

\bibitem{bai2018empirical}
\bibinfo{author}{Bai, S.}, \bibinfo{author}{Kolter, J.~Z.} \& \bibinfo{author}{Koltun, V.}
\newblock \bibinfo{journal}{\bibinfo{title}{An empirical evaluation of generic convolutional and recurrent networks for sequence modeling}}.
\newblock {\emph{\JournalTitle{arXiv preprint arXiv:1803.01271}}}  (\bibinfo{year}{2018}).

\bibitem{lea2017temporal}
\bibinfo{author}{Lea, C.}, \bibinfo{author}{Flynn, M.~D.}, \bibinfo{author}{Vidal, R.}, \bibinfo{author}{Reiter, A.} \& \bibinfo{author}{Hager, G.~D.}
\newblock \bibinfo{title}{Temporal convolutional networks for action segmentation and detection}.
\newblock In \emph{\bibinfo{booktitle}{proceedings of the IEEE Conference on Computer Vision and Pattern Recognition}}, \bibinfo{pages}{156--165} (\bibinfo{year}{2017}).

\bibitem{franceschi2019unsupervised}
\bibinfo{author}{Franceschi, J.-Y.}, \bibinfo{author}{Dieuleveut, A.} \& \bibinfo{author}{Jaggi, M.}
\newblock \bibinfo{journal}{\bibinfo{title}{Unsupervised scalable representation learning for multivariate time series}}.
\newblock {\emph{\JournalTitle{Advances in neural information processing systems}}} \textbf{\bibinfo{volume}{32}} (\bibinfo{year}{2019}).

\bibitem{ismail2019deep}
\bibinfo{author}{Ismail~Fawaz, H.}, \bibinfo{author}{Forestier, G.}, \bibinfo{author}{Weber, J.}, \bibinfo{author}{Idoumghar, L.} \& \bibinfo{author}{Muller, P.-A.}
\newblock \bibinfo{journal}{\bibinfo{title}{Deep learning for time series classification: a review}}.
\newblock {\emph{\JournalTitle{Data mining and knowledge discovery}}} \textbf{\bibinfo{volume}{33}}, \bibinfo{pages}{917--963} (\bibinfo{year}{2019}).

\bibitem{bagnall2017great}
\bibinfo{author}{Bagnall, A.}, \bibinfo{author}{Lines, J.}, \bibinfo{author}{Bostrom, A.}, \bibinfo{author}{Large, J.} \& \bibinfo{author}{Keogh, E.}
\newblock \bibinfo{journal}{\bibinfo{title}{The great time series classification bake off: a review and experimental evaluation of recent algorithmic advances}}.
\newblock {\emph{\JournalTitle{Data mining and knowledge discovery}}} \textbf{\bibinfo{volume}{31}}, \bibinfo{pages}{606--660} (\bibinfo{year}{2017}).

\bibitem{daley2006running}
\bibinfo{author}{Daley, M.~A.} \& \bibinfo{author}{Biewener, A.~A.}
\newblock \bibinfo{journal}{\bibinfo{title}{Running over rough terrain reveals limb control for intrinsic stability}}.
\newblock {\emph{\JournalTitle{Proceedings of the National Academy of Sciences}}} \textbf{\bibinfo{volume}{103}}, \bibinfo{pages}{15681--15686} (\bibinfo{year}{2006}).

\bibitem{van2007muscle}
\bibinfo{author}{van~der Linden, M.~H.}, \bibinfo{author}{Marigold, D.~S.}, \bibinfo{author}{Gabre{\"e}ls, F.~J.} \& \bibinfo{author}{Duysens, J.}
\newblock \bibinfo{journal}{\bibinfo{title}{Muscle reflexes and synergies triggered by an unexpected support surface height during walking}}.
\newblock {\emph{\JournalTitle{Journal of neurophysiology}}} \textbf{\bibinfo{volume}{97}}, \bibinfo{pages}{3639--3650} (\bibinfo{year}{2007}).

\bibitem{merel2019hierarchical}
\bibinfo{author}{Merel, J.}, \bibinfo{author}{Botvinick, M.} \& \bibinfo{author}{Wayne, G.}
\newblock \bibinfo{journal}{\bibinfo{title}{Hierarchical motor control in mammals and machines}}.
\newblock {\emph{\JournalTitle{Nature communications}}} \textbf{\bibinfo{volume}{10}}, \bibinfo{pages}{5489} (\bibinfo{year}{2019}).

\bibitem{kuo2002relative}
\bibinfo{author}{Kuo, A.~D.}
\newblock \bibinfo{journal}{\bibinfo{title}{The relative roles of feedforward and feedback in the control of rhythmic movements}}.
\newblock {\emph{\JournalTitle{Motor control}}} \textbf{\bibinfo{volume}{6}}, \bibinfo{pages}{129--145} (\bibinfo{year}{2002}).

\bibitem{todorov2002optimal}
\bibinfo{author}{Todorov, E.} \& \bibinfo{author}{Jordan, M.~I.}
\newblock \bibinfo{journal}{\bibinfo{title}{Optimal feedback control as a theory of motor coordination}}.
\newblock {\emph{\JournalTitle{Nature neuroscience}}} \textbf{\bibinfo{volume}{5}}, \bibinfo{pages}{1226--1235} (\bibinfo{year}{2002}).

\bibitem{maurus2024increased}
\bibinfo{author}{Maurus, P.}, \bibinfo{author}{Mahdi, G.} \& \bibinfo{author}{Cluff, T.}
\newblock \bibinfo{journal}{\bibinfo{title}{Increased muscle coactivation is linked with fast feedback control when reaching in unpredictable visual environments}}.
\newblock {\emph{\JournalTitle{iScience}}} \textbf{\bibinfo{volume}{27}} (\bibinfo{year}{2024}).

\bibitem{nayeem2021preparing}
\bibinfo{author}{Nayeem, R.}, \bibinfo{author}{Bazzi, S.}, \bibinfo{author}{Sadeghi, M.}, \bibinfo{author}{Hogan, N.} \& \bibinfo{author}{Sternad, D.}
\newblock \bibinfo{journal}{\bibinfo{title}{Preparing to move: Setting initial conditions to simplify interactions with complex objects}}.
\newblock {\emph{\JournalTitle{PLOS Computational Biology}}} \textbf{\bibinfo{volume}{17}}, \bibinfo{pages}{e1009597} (\bibinfo{year}{2021}).

\bibitem{leestma2023linking}
\bibinfo{author}{Leestma, J.~K.}, \bibinfo{author}{Golyski, P.~R.}, \bibinfo{author}{Smith, C.~R.}, \bibinfo{author}{Sawicki, G.~S.} \& \bibinfo{author}{Young, A.~J.}
\newblock \bibinfo{journal}{\bibinfo{title}{Linking whole-body angular momentum and step placement during perturbed human walking}}.
\newblock {\emph{\JournalTitle{Journal of Experimental Biology}}} \textbf{\bibinfo{volume}{226}}, \bibinfo{pages}{jeb244760} (\bibinfo{year}{2023}).

\bibitem{van2025simultaneous}
\bibinfo{author}{van Dieen, J.~H.}, \bibinfo{author}{Bruijn, S.~M.}, \bibinfo{author}{Lemaire, K.~K.} \& \bibinfo{author}{Kistemaker, D.~A.}
\newblock \bibinfo{journal}{\bibinfo{title}{Simultaneous stabilizing feedback control of linear and angular momentum in human walking}}.
\newblock {\emph{\JournalTitle{bioRxiv}}} \bibinfo{pages}{2025--01} (\bibinfo{year}{2025}).

\bibitem{maclellan2006adaptations}
\bibinfo{author}{MacLellan, M.~J.} \& \bibinfo{author}{Patla, A.~E.}
\newblock \bibinfo{journal}{\bibinfo{title}{Adaptations of walking pattern on a compliant surface to regulate dynamic stability}}.
\newblock {\emph{\JournalTitle{Experimental brain research}}} \textbf{\bibinfo{volume}{173}}, \bibinfo{pages}{521--530} (\bibinfo{year}{2006}).

\bibitem{hak2013steps}
\bibinfo{author}{Hak, L.}, \bibinfo{author}{Houdijk, H.}, \bibinfo{author}{Beek, P.~J.} \& \bibinfo{author}{van Die{\"e}n, J.~H.}
\newblock \bibinfo{journal}{\bibinfo{title}{Steps to take to enhance gait stability: the effect of stride frequency, stride length, and walking speed on local dynamic stability and margins of stability}}.
\newblock {\emph{\JournalTitle{PloS one}}} \textbf{\bibinfo{volume}{8}}, \bibinfo{pages}{e82842} (\bibinfo{year}{2013}).

\bibitem{schulz2012healthy}
\bibinfo{author}{Schulz, B.~W.}
\newblock \bibinfo{journal}{\bibinfo{title}{Healthy younger and older adults control foot placement to avoid small obstacles during gait primarily by modulating step width}}.
\newblock {\emph{\JournalTitle{Journal of neuroengineering and rehabilitation}}} \textbf{\bibinfo{volume}{9}}, \bibinfo{pages}{1--10} (\bibinfo{year}{2012}).

\bibitem{collins2013two}
\bibinfo{author}{Collins, S.~H.} \& \bibinfo{author}{Kuo, A.~D.}
\newblock \bibinfo{journal}{\bibinfo{title}{Two independent contributions to step variability during over-ground human walking}}.
\newblock {\emph{\JournalTitle{PloS one}}} \textbf{\bibinfo{volume}{8}}, \bibinfo{pages}{e73597} (\bibinfo{year}{2013}).

\bibitem{taheri2020grab}
\bibinfo{author}{Taheri, O.}, \bibinfo{author}{Ghorbani, N.}, \bibinfo{author}{Black, M.~J.} \& \bibinfo{author}{Tzionas, D.}
\newblock \bibinfo{title}{Grab: A dataset of whole-body human grasping of objects}.
\newblock In \emph{\bibinfo{booktitle}{Computer Vision--ECCV 2020: 16th European Conference, Glasgow, UK, August 23--28, 2020, Proceedings, Part IV 16}}, \bibinfo{pages}{581--600} (\bibinfo{organization}{Springer}, \bibinfo{year}{2020}).

\bibitem{kratzer2020mogaze}
\bibinfo{author}{Kratzer, P.} \emph{et~al.}
\newblock \bibinfo{journal}{\bibinfo{title}{Mogaze: A dataset of full-body motions that includes workspace geometry and eye-gaze}}.
\newblock {\emph{\JournalTitle{IEEE Robotics and Automation Letters}}} \textbf{\bibinfo{volume}{6}}, \bibinfo{pages}{367--373} (\bibinfo{year}{2020}).

\bibitem{schwaner2021future}
\bibinfo{author}{Schwaner, M.} \emph{et~al.}
\newblock \bibinfo{journal}{\bibinfo{title}{Future tail tales: A forward-looking, integrative perspective on tail research}}.
\newblock {\emph{\JournalTitle{Integrative and Comparative Biology}}} \textbf{\bibinfo{volume}{61}}, \bibinfo{pages}{521--537} (\bibinfo{year}{2021}).

\bibitem{muramatsu2025wildpose}
\bibinfo{author}{Muramatsu, N.}, \bibinfo{author}{Shin, S.}, \bibinfo{author}{Deng, Q.}, \bibinfo{author}{Markham, A.} \& \bibinfo{author}{Patel, A.}
\newblock \bibinfo{journal}{\bibinfo{title}{Wildpose: A long-range 3d wildlife motion capture system}}.
\newblock {\emph{\JournalTitle{Journal of Experimental Biology}}} \bibinfo{pages}{jeb--249987} (\bibinfo{year}{2025}).

\bibitem{song2015neural}
\bibinfo{author}{Song, S.} \& \bibinfo{author}{Geyer, H.}
\newblock \bibinfo{journal}{\bibinfo{title}{A neural circuitry that emphasizes spinal feedback generates diverse behaviours of human locomotion}}.
\newblock {\emph{\JournalTitle{The Journal of physiology}}} \textbf{\bibinfo{volume}{593}}, \bibinfo{pages}{3493--3511} (\bibinfo{year}{2015}).

\bibitem{song2017evaluation}
\bibinfo{author}{Song, S.} \& \bibinfo{author}{Geyer, H.}
\newblock \bibinfo{journal}{\bibinfo{title}{Evaluation of a neuromechanical walking control model using disturbance experiments}}.
\newblock {\emph{\JournalTitle{Frontiers in computational neuroscience}}} \textbf{\bibinfo{volume}{11}}, \bibinfo{pages}{15} (\bibinfo{year}{2017}).

\bibitem{caggiano2022myosuite}
\bibinfo{author}{Caggiano, V.}, \bibinfo{author}{Wang, H.}, \bibinfo{author}{Durandau, G.}, \bibinfo{author}{Sartori, M.} \& \bibinfo{author}{Kumar, V.}
\newblock \bibinfo{title}{Myosuite: A contact-rich simulation suite for musculoskeletal motor control}.
\newblock In \emph{\bibinfo{booktitle}{Learning for Dynamics and Control Conference}}, \bibinfo{pages}{492--507} (\bibinfo{organization}{PMLR}, \bibinfo{year}{2022}).

\bibitem{giggins2013biofeedback}
\bibinfo{author}{Giggins, O.~M.}, \bibinfo{author}{Persson, U.~M.} \& \bibinfo{author}{Caulfield, B.}
\newblock \bibinfo{journal}{\bibinfo{title}{Biofeedback in rehabilitation}}.
\newblock {\emph{\JournalTitle{Journal of neuroengineering and rehabilitation}}} \textbf{\bibinfo{volume}{10}}, \bibinfo{pages}{60} (\bibinfo{year}{2013}).

\bibitem{hribernik2022review}
\bibinfo{author}{Hribernik, M.}, \bibinfo{author}{Umek, A.}, \bibinfo{author}{Toma{\v{z}}i{\v{c}}, S.} \& \bibinfo{author}{Kos, A.}
\newblock \bibinfo{journal}{\bibinfo{title}{Review of real-time biomechanical feedback systems in sport and rehabilitation}}.
\newblock {\emph{\JournalTitle{Sensors}}} \textbf{\bibinfo{volume}{22}}, \bibinfo{pages}{3006} (\bibinfo{year}{2022}).

\bibitem{bao2020vision}
\bibinfo{author}{Bao, W.}, \bibinfo{author}{Villarreal, D.} \& \bibinfo{author}{Chiao, J.-C.}
\newblock \bibinfo{title}{Vision-based autonomous walking in a lower-limb powered exoskeleton}.
\newblock In \emph{\bibinfo{booktitle}{2020 IEEE 20th International Conference on Bioinformatics and Bioengineering (BIBE)}}, \bibinfo{pages}{830--834} (\bibinfo{organization}{IEEE}, \bibinfo{year}{2020}).

\bibitem{kang2019real}
\bibinfo{author}{Kang, I.}, \bibinfo{author}{Kunapuli, P.} \& \bibinfo{author}{Young, A.~J.}
\newblock \bibinfo{journal}{\bibinfo{title}{Real-time neural network-based gait phase estimation using a robotic hip exoskeleton}}.
\newblock {\emph{\JournalTitle{IEEE Transactions on Medical Robotics and Bionics}}} \textbf{\bibinfo{volume}{2}}, \bibinfo{pages}{28--37} (\bibinfo{year}{2019}).

\bibitem{banks2015using}
\bibinfo{author}{Banks, J.~J.}, \bibinfo{author}{Chang, W.-R.}, \bibinfo{author}{Xu, X.} \& \bibinfo{author}{Chang, C.-C.}
\newblock \bibinfo{journal}{\bibinfo{title}{Using horizontal heel displacement to identify heel strike instants in normal gait}}.
\newblock {\emph{\JournalTitle{Gait \& Posture}}} \textbf{\bibinfo{volume}{42}}, \bibinfo{pages}{101--103} (\bibinfo{year}{2015}).

\bibitem{zeni2008two}
\bibinfo{author}{Zeni~Jr, J.}, \bibinfo{author}{Richards, J.} \& \bibinfo{author}{Higginson, J.}
\newblock \bibinfo{journal}{\bibinfo{title}{Two simple methods for determining gait events during treadmill and overground walking using kinematic data}}.
\newblock {\emph{\JournalTitle{Gait \& posture}}} \textbf{\bibinfo{volume}{27}}, \bibinfo{pages}{710--714} (\bibinfo{year}{2008}).

\bibitem{hestness2017deep}
\bibinfo{author}{Hestness, J.} \emph{et~al.}
\newblock \bibinfo{journal}{\bibinfo{title}{Deep learning scaling is predictable, empirically}}.
\newblock {\emph{\JournalTitle{arXiv preprint arXiv:1712.00409}}}  (\bibinfo{year}{2017}).

\bibitem{guo2016entity}
\bibinfo{author}{Guo, C.} \& \bibinfo{author}{Berkhahn, F.}
\newblock \bibinfo{journal}{\bibinfo{title}{Entity embeddings of categorical variables}}.
\newblock {\emph{\JournalTitle{arXiv preprint arXiv:1604.06737}}}  (\bibinfo{year}{2016}).

\bibitem{hochreiter1997long}
\bibinfo{author}{Hochreiter, S.} \& \bibinfo{author}{Schmidhuber, J.}
\newblock \bibinfo{journal}{\bibinfo{title}{Long short-term memory}}.
\newblock {\emph{\JournalTitle{Neural computation}}} \textbf{\bibinfo{volume}{9}}, \bibinfo{pages}{1735--1780} (\bibinfo{year}{1997}).

\bibitem{karpathy2015deep}
\bibinfo{author}{Karpathy, A.} \& \bibinfo{author}{Fei-Fei, L.}
\newblock \bibinfo{title}{Deep visual-semantic alignments for generating image descriptions}.
\newblock In \emph{\bibinfo{booktitle}{Proceedings of the IEEE conference on computer vision and pattern recognition}}, \bibinfo{pages}{3128--3137} (\bibinfo{year}{2015}).

\bibitem{bates2024cross}
\bibinfo{author}{Bates, S.}, \bibinfo{author}{Hastie, T.} \& \bibinfo{author}{Tibshirani, R.}
\newblock \bibinfo{journal}{\bibinfo{title}{Cross-validation: what does it estimate and how well does it do it?}}
\newblock {\emph{\JournalTitle{Journal of the American Statistical Association}}} \textbf{\bibinfo{volume}{119}}, \bibinfo{pages}{1434--1445} (\bibinfo{year}{2024}).

\bibitem{kingma2014adam}
\bibinfo{author}{Kingma, D.~P.} \& \bibinfo{author}{Ba, J.}
\newblock \bibinfo{journal}{\bibinfo{title}{Adam: A method for stochastic optimization}}.
\newblock {\emph{\JournalTitle{arXiv preprint arXiv:1412.6980}}}  (\bibinfo{year}{2014}).

\bibitem{cleveland1979robust}
\bibinfo{author}{Cleveland, W.~S.}
\newblock \bibinfo{journal}{\bibinfo{title}{Robust locally weighted regression and smoothing scatterplots}}.
\newblock {\emph{\JournalTitle{Journal of the American statistical association}}} \textbf{\bibinfo{volume}{74}}, \bibinfo{pages}{829--836} (\bibinfo{year}{1979}).

\bibitem{takens2006detecting}
\bibinfo{author}{Takens, F.}
\newblock \bibinfo{title}{Detecting strange attractors in turbulence}.
\newblock In \emph{\bibinfo{booktitle}{Dynamical Systems and Turbulence, Warwick 1980: proceedings of a symposium held at the University of Warwick 1979/80}}, \bibinfo{pages}{366--381} (\bibinfo{organization}{Springer}, \bibinfo{year}{2006}).

\bibitem{yang2020hyperparameter}
\bibinfo{author}{Yang, L.} \& \bibinfo{author}{Shami, A.}
\newblock \bibinfo{journal}{\bibinfo{title}{On hyperparameter optimization of machine learning algorithms: Theory and practice}}.
\newblock {\emph{\JournalTitle{Neurocomputing}}} \textbf{\bibinfo{volume}{415}}, \bibinfo{pages}{295--316} (\bibinfo{year}{2020}).

\bibitem{cawley2010over}
\bibinfo{author}{Cawley, G.~C.} \& \bibinfo{author}{Talbot, N.~L.}
\newblock \bibinfo{journal}{\bibinfo{title}{On over-fitting in model selection and subsequent selection bias in performance evaluation}}.
\newblock {\emph{\JournalTitle{The Journal of Machine Learning Research}}} \textbf{\bibinfo{volume}{11}}, \bibinfo{pages}{2079--2107} (\bibinfo{year}{2010}).

\bibitem{bergstra2012random}
\bibinfo{author}{Bergstra, J.} \& \bibinfo{author}{Bengio, Y.}
\newblock \bibinfo{journal}{\bibinfo{title}{Random search for hyper-parameter optimization.}}
\newblock {\emph{\JournalTitle{Journal of machine learning research}}} \textbf{\bibinfo{volume}{13}} (\bibinfo{year}{2012}).

\end{thebibliography}
